%% file: main.tex
\documentclass[11pt]{eccc}
\usepackage{fullpage}
\title[Learning in the Presence of Arbitrary Outliers]{Resilience: A Criterion for Learning in the Presence of Arbitrary Outliers}
\usepackage{times}
\usepackage{bbm}
\usepackage{algorithm}
\usepackage{algorithmic,algorithm}
 \coltauthor{\Name{Jacob Steinhardt}\thanks{JS was supported by a Fannie \& John Hertz Foundation Fellowship, an NSF Graduate Research Fellowship, and a Future of Life Instute grant.} \Email{jsteinha@stanford.edu}\\
 \Name{Moses Charikar}\thanks{MC was supported by NSF grants CCF-1617577, CCF-1302518 and a Simons Investigator Award.} \Email{moses@stanford.edu}\\
 \Name{Gregory Valiant}\thanks{GV was supported by NSF CAREER Award CCF-1351108 and a Sloan Research Fellowship.} \Email{valiant@stanford.edu}\\
 }

\input latex-defs
\input macros

\begin{document}

\maketitle

\vskip -0.3in
\input abstract
\input introduction

\input info-theory

\input powering-up
\input main-lp
\input rank-k

\input concentration

\input info-theory-applications

\input algo-applications

\appendix

\input counterexample
\input resilience-properties
\input 1st-moment-lp-proof
\input minimax-proof
\input powering-up-lp-eps-proof
\input unitary-proof
\input invariant-proof
\input concentration-cov-proof

\bibliography{refdb/all}

\end{document}

%% file: latex-defs.tex
\def\[#1\]{\begin{align}#1\end{align}}
\def\(#1\){\begin{align*}#1\end{align*}}

\DeclareMathOperator{\col}{col}
\DeclareMathOperator{\rank}{rank}
\newcommand{\minim}[1]{\underset{#1}{\text{minimize}}}
\newcommand{\subjto}{\text{subject to}}
\newcommand{\opsep}{\phantom{+}}
\newcommand{\fasep}{\,\,}

\newcommand{\oo}{\mathcal{O}}
\newcommand{\oot}{\widetilde{\mathcal{O}}}
\newcommand{\p}[1]{\left(#1\right)}

\DeclareMathOperator{\diag}{diag}

\DeclareMathOperator{\tr}{tr}
\newcommand{\eqdef}{\stackrel{\mathrm{def}}{=}}
\newcommand{\bi}{\mathbbm{1}}
\newcommand{\bP}{\mathbb{P}}
\newcommand{\bI}{\mathbb{I}}
\newcommand{\bE}{\mathbb{E}}
\newcommand{\sV}{\mathcal{V}}

\newcommand{\sA}{\mathcal{A}}

\newcommand{\bR}{\mathbb{R}}

\newcommand{\sP}{\mathcal{P}}
\newtheorem{assumption}[theorem]{Assumption}
\newtheorem*{theorem*}{Theorem}
\newtheorem*{proposition*}{Proposition}

%% file: macros.tex
\newcommand{\St}{\tilde{S}}

\newcommand{\sigmabest}{\sigma_*}
\newcommand{\sigmastar}{\widetilde{\sigma}_*}
\newcommand{\qed}{\jmlrQED}
\renewcommand{\paragraph}[1]{\vskip 1pt plus1pt \noindent \textbf{#1}}

\newcommand{\TV}{TV}

%% file: abstract.tex
\begin{abstract}
We introduce a criterion, \emph{resilience}, which allows properties of a dataset 
(such as its mean or best low rank approximation) to be robustly computed, even 
in the presence of a large fraction of arbitrary additional data. Resilience is a 
weaker condition than most other properties considered so far in the literature, 
and yet enables robust estimation in a broader variety of settings. We provide new 
information-theoretic results on robust distribution learning, robust estimation of 
stochastic block models, and robust mean estimation under bounded $k$th moments. 
We also provide new algorithmic results on robust distribution learning, as well as 
robust mean estimation in $\ell_p$-norms. Among our proof techniques is a method for 
pruning a high-dimensional distribution with bounded $1$st moments to a stable ``core'' 
with bounded $2$nd moments, which may be of independent interest.
\end{abstract}


%% file: introduction.tex
\section{Introduction}

What are the fundamental properties that allow one to robustly learn from a dataset, even 
if some fraction of that dataset consists of arbitrarily corrupted data? While much work 
has been done in the setting of noisy data, or for restricted families of outliers, 
it is only recently 
that provable algorithms for learning in the presence of a large fraction of arbitrary 
(and potentially adversarial) data have been formulated in high-dimensional settings 
\citep{klivans2009learning,xu2010principal,diakonikolas2016robust,lai2016agnostic,
steinhardt2016avoiding,charikar2017learning}.
In this work, we formulate a conceptually simple criterion that a dataset can 
satisfy--\emph{resilience}--which guarantees that properties such as the mean of that dataset 
can be robustly estimated even if a large fraction of additional arbitrary data is 
inserted.

To illustrate our setting, consider the following game between Alice (the adversary) and Bob. 
First, a set $S \subseteq \bR^d$ of $(1-\epsilon)n$ points is given to Alice. Alice then adds $\epsilon n$ 
additional points to $S$ to create a new set $\St$, and passes $\St$ to Bob. Bob 
wishes to output a parameter $\hat{\mu}$ that is as close as possible to the mean $\mu$ of 
the points in the original set $S$, with error measured according to some norm $\|\hat{\mu} - \mu\|$. 
The question is: how well can Bob do, assuming that Alice is an adversary with knowledge of Bob's algorithm?

\begin{figure}
\begin{center}
%\includegraphics[width=0.3\textwidth]{initial.png}
%\hspace{1em}
%\includegraphics[width=0.3\textwidth]{alice.png}
%\hspace{1em}
\includegraphics[width=0.28\textwidth]{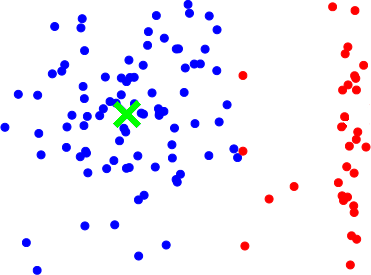}
\end{center}
\caption{Illustration of the robust mean estimation setting. 
First, a set of points (blue) is given to Alice, who adds an $\epsilon$ 
fraction of adversarially chosen points (red). Bob's goal is to 
output the mean of the original set (indicated in green).}
\label{fig:setting}
\end{figure}

The above game models mean estimation in the presence of arbitrary outliers; one can easily consider 
other problems as well (e.g. regression) but we will focus on mean estimation in this paper.

With no asumptions on $S$, Bob will clearly incur arbitrarily large error in the worst case---Alice 
can add points arbitrarily far away from the true mean $\mu$, and Bob has no way of telling whether 
those points actually belong to $S$ or were added by Alice. A first pass assumption is to suppose that 
$S$ has diameter at most $\rho$; then by discarding points that are very far away from most other points, 
Bob can obtain error $\oo(\epsilon \rho)$. However, in most high-dimensional settings, the diameter 
$\rho$ grows polynomially with the dimension $d$ (e.g. the $d$-dimensional hypercube has $\ell_2$-diameter 
$\Theta(\sqrt{d})$). Subtler criteria are therefore needed to obtain dimension-independent bounds in 
most settings of interest.

Recently, \citet{diakonikolas2016robust} showed that Bob can incur $\ell_2$-error 
$\oo(\epsilon\sqrt{\log(1/\epsilon)})$ when the points in $S$ are drawn from a $d$-dimensional Gaussian, 
while \citet{lai2016agnostic} concurrently showed that Bob can incur $\ell_2$-error $\oo(\sqrt{\epsilon \log(d)})$ 
if the points in $S$ are drawn from a distribution with bounded $4$th moments. Since then, a considerable 
amount of additional work has studied high-dimensional estimation in the presence of adversaries, which 
we discuss in detail below. However, in general, both Bob's strategy and its analysis tend to be quite 
complex, and specialized to particular distributional assumptions. This raises the question---is it 
possible to formulate a general and simple-to-understand criterion for the set $S$ under which Bob has a 
(possibly inefficient) strategy for incurring small error?

In this paper, we provide such a criterion; we identify an assumption--\emph{resilience}--on the set $S$, 
under which Bob has a straightforward exponential-time algorithm for estimating $\mu$ accurately. This yields new 
information-theoretic bounds for a number of robust learning problems, including robust learning of stochastic 
block models, of discrete distributions, and of distributions with bounded $k$th moments. 
We also identify additional 
assumptions under which Bob has an efficient (polynomial-time) strategy for estimating $\mu$, 
which yields an efficient algorithm for robust learning of discrete distributions, 
as well as for robust mean estimation in $\ell_p$-norms.

The resilience condition is essentially that the mean of every large subset of $S$ must be 
close to the mean of all of $S$. More formally, for a norm $\|\cdot\|$, our criterion is as follows:
\begin{definition}[Resilience]
\label{def:resilience-intro}
\label{def:resilience-lp}
A set of points $\{x_i\}_{i \in S}$ lying in $\bR^d$ is \emph{$(\sigma,\epsilon)$-resilient} 
in a norm 
$\|\cdot\|$ around a point $\mu$ if, for all subsets $T \subseteq S$ of size at least 
$(1-\epsilon)|S|$,
\begin{equation}
\label{eq:resilience-intro}
\Big\|\frac{1}{|T|} \sum_{i \in T} (x_i - \mu)\Big\| \leq \sigma.
\end{equation}
%Given a norm $\|\cdot\|$, 
More generally, a distribution $p$ is said to be $(\sigma,\epsilon)$-resilient 
if $\|\bE[x - \mu \mid E]\| \leq \sigma$ for every event $E$ of probability 
at least $1-\epsilon$.
\end{definition}
In the definition above,
$\mu$ need not equal the mean of $S$; this distinction is useful 
in statistical settings 
where the sample mean of a finite set of points differs slightly from the true mean. 
However, \eqref{eq:resilience-intro} implies that $\mu$ differs from the mean of 
$S$ by at most $\sigma$.
%However, by setting $T = S$ in \eqref{eq:resilience-intro} we see that $\mu$ 
%differs from the mean of $S$ by at most $\sigma$.

Importantly, Definition~\ref{def:resilience-intro} is satisfied with high probability by a 
finite sample in many settings. For instance, samples from a distribution with $k$th moments 
bounded by $\sigma$ will be $(\oo(\sigma \epsilon^{1-1/k}), \epsilon)$-resilient in the 
$\ell_2$-norm with high probability. Resilience also holds with high probability under many 
other natural distributional assumptions, discussed in more detail in 
Sections~\ref{sec:info-theory-intro}, \ref{sec:concentration}, 
and \ref{sec:info-theory-applications}.

Assuming that the original set $S$ is $(\sigma,\epsilon)$-resilient, 
Bob's strategy is actually quite simple---find \emph{any} large 
$(\sigma,\epsilon)$-resilient subset $S'$ of the corrupted set $\St$, 
and output the mean of $S'$. By pigeonhole, $S'$ and $S$ have large intersection, and 
hence by \eqref{eq:resilience-intro} must have similar means. 
We establish this formally in Section~\ref{sec:info-theory}.

Pleasingly, resilience reduces the question of whether Bob can win the game to a 
purely algorithmic question---that of finding any large resilient set. Rather than 
wondering whether it is even information-theoretically possible to estimate $\mu$, we 
can instead focus on efficiently finding resilient subsets of $\St$.

We provide one such algorithm in Section~\ref{sec:alg}, assuming that the norm $\|\cdot\|$
is \emph{strongly convex} and that we can approximately solve a certain generalized 
eigenvalue problem in the dual norm. When specialized 
to the $\ell_1$-norm, our general algorithm yields an efficient procedure for robust learning of discrete 
distributions.

In the remainder of this section, we will outline our main results, starting with 
information-theoretic results and then moving on to algorithmic results. 
In Section~\ref{sec:info-theory-intro}, we show that resilience is indeed 
information-theoretically sufficient for robust mean estimation. In 
Section~\ref{sec:intro-concentration}, we then provide finite-sample bounds showing that 
resilience holds with high probability for i.i.d.~samples from a distribution. 

In Section~\ref{sec:algo-intro}, we turn our attention to algorithmic bounds. 
We identity a property--bounded variance in the dual norm--under which efficient 
algorithms exist. We then show that, as long as the norm is {strongly convex}, every 
resilient set has a large subset with bounded variance, thus enabling efficient 
algorithms. This connection between 
resilience and bounded variance is the most technically non-trivial component of our 
results, and may be of independent interest. 

Both our information-theoretic and algorithmic 
bounds yield new results in concrete settings, which we discuss in the corresponding subsections.
In Section~\ref{sec:rank-intro}, we also
discuss an extension of resilience to low-rank matrix approximation, 
which enables us to derive new bounds in that setting as well.
In Section~\ref{sec:roadmap} we outline the rest of the paper and point to technical 
highlights, and in Section~\ref{sec:related-work} we discuss related work.

\subsection{Information-Theoretic Sufficiency}
\label{sec:info-theory-intro}

First, we show that resilience is indeed information-theoretically 
sufficient for robust recovery of the mean $\mu$. In what follows, we use 
$\sigmabest(\epsilon)$ to denote the smallest $\sigma$ such that $S$ is 
$(\sigma,\epsilon)$-resilient.
\begin{proposition}
\label{prop:resilience}
Suppose that $\St = \{x_1,\ldots,x_n\}$ contains a set $S$ of size $(1-\epsilon)n$ 
that is resilient around $\mu$ (where $S$ and $\mu$ are both unknown). 
Then if $\epsilon < \frac{1}{2}$, it is possible to recover a 
$\hat{\mu}$ such that $\|\hat{\mu} - \mu\| \leq 2\sigmabest(\frac{\epsilon}{1-\epsilon})$.

More generally, if $|S| \geq \alpha n$ (even if $\alpha < \frac{1}{2}$), 
it is possible to output a (random) $\hat{\mu}$ such that 
$\|\hat{\mu} - \mu\| \leq \frac{16}{\alpha}\sigmabest(\frac{\alpha}{4})$ 
with probability at least $\frac{\alpha}{2}$.
\end{proposition}

The first part says that robustness to an $\epsilon$ fraction of outliers 
%(i.e., if $\alpha = 1-\epsilon$) 
depends on resilience to a $\frac{\epsilon}{1-\epsilon}$ fraction 
of deletions. Thus, Bob has a good strategy as long as $\sigmabest(\frac{\epsilon}{1-\epsilon})$ 
is small.

The second part, which is more surprising, says that Bob has a good strategy 
\emph{even if the majority of $\St$ is controlled by Alice}. Here one cannot hope for recovery 
in the usual sense, because if $\alpha = \frac{1}{2}$ (i.e., Alice controls half the points) 
then Alice can make $\St$ the disjoint union of two identical copies of $S$ (one of which is 
shifted by a large amount) and Bob has no way of determining which of the two copies is the 
true $S$. Nevertheless, in this situation Bob can still identify $S$ (and hence $\mu$) with 
probability $\frac{1}{2}$; more generally, the second part of Proposition~\ref{prop:resilience} 
says that if $|S| = \alpha |\St|$ then Bob can identify $\mu$ with probability at least 
$\frac{\alpha}{2}$. 
%In this case, we require $S$ to be resilient even if we remove all but an 
%$\frac{\alpha}{2}$ fraction of the points.

The fact that estimation is possible even when $\alpha < \frac{1}{2}$ was first established by 
\citet{steinhardt2016avoiding} in a crowdsourcing setting, and later by 
\citet{charikar2017learning} in a number of settings including mean estimation.
Apart from being interesting due to its unexpectedness, estimation in this regime has 
immediate implications for robust estimation of mixtures of distributions (by considering each 
mixture component in turn as the ``good'' set $S$) or of planted substructures in random graphs.
We refer the reader to \citet{charikar2017learning} for a full elaboration of this point.

The proof of Proposition~\ref{prop:resilience}, given in detail in 
Section~\ref{sec:resilience-proof}, is a pigeonhole argument. For the 
$\epsilon < \frac{1}{2}$ case, we simply search for any large resilient set $S'$ and output 
its mean; then $S$ and $S'$ must have large overlap, and by resilience their means 
must both be close to the mean of their intersection, and hence to each other.

For the general case where $|S| = \alpha |\St|$ (possibly with $\alpha < \frac{1}{2}$), 
a similar pigeonhole argument applies but we now need to consider a covering of 
$\St$ by $\frac{2}{\alpha}$ approximately disjoint sets $S_1', \ldots, S_{2/\alpha}'$. We 
can show that the true set $S$ must overlap at least one of these sets by a decent amount, 
and so outputting the mean of one of these sets at random gives a good approximation 
to the mean of $S$ with probability $\frac{\alpha}{2}$.
%\todo{outline Pigeonhole argument?}

\subsection{Finite-Sample Concentration}
\label{sec:intro-concentration}

While Proposition~\ref{prop:resilience} provides a deterministic condition under 
which robust mean estimation is possible, we would also like a way of checking that 
resilience holds with high probability given samples $x_1, \ldots, x_n$ from a distribution 
$p$. First, we provide an alternate characterization of resilience which says that a 
distribution is resilient if it has \emph{thin tails} in every direction:
\begin{lemma}
\label{lem:tail-bound}
Given a norm $\|\cdot\|$, define the dual norm 
$\|v\|_* = \sup_{\|x\| \leq 1} \langle v, x \rangle$. 
For a fixed vector $v$, let $\tau_{\epsilon}(v)$ denote the 
$\epsilon$-quantile of $\langle x - \mu, v \rangle$: 
$\bP_{x \sim p}[\langle x - \mu, v \rangle \geq \tau_{\epsilon}(v)] = \epsilon$. 
Then, $p$ is $(\sigma,\epsilon)$-resilient around its mean $\mu$ if and only if 
\begin{equation}
\bE_p[\langle x - \mu, v \rangle \mid \langle x - \mu, v \rangle \geq \tau_{\epsilon}(v)] \leq \frac{1-\epsilon}{\epsilon}\sigma \text{ whenever } \|v\|_* \leq 1.
\end{equation}
%whenever $\|v\|_* \leq 1$. \todo{save line}
\end{lemma}
In other words, if we project onto any unit vector $v$ in the dual norm, 
the $\epsilon$-tail of $x-\mu$ must have mean at most $\frac{1-\epsilon}{\epsilon}\sigma$.
Thus, for instance, a distribution with variance at most $\sigma_0^2$ along every unit vector 
would have $\sigma = \oo(\sigma_0 \sqrt{\epsilon})$.
Note that Lemma~\ref{lem:tail-bound} requires $\mu$ to be the mean, 
rather than an arbitrary vector as before.

We next provide a meta-result establishing that resilience of a 
population distribution $p$ very generically transfers to a finite set of samples from that 
distribution. The number of samples necessary depends on two quantities $B$ and $\log M$ 
that will be defined in detail later; for now we note that they are ways of measuring 
the effective dimension of the space.
\begin{proposition}
\label{prop:concentration-resilient}
Suppose that a distribution $p$ is $(\sigma,\epsilon)$-resilient around its mean $\mu$ with $\epsilon < \frac{1}{2}$.
%$\|\bE[x \mid E] - \mu\| \leq \sigma$ for any event 
%$E$ of probability at least $\epsilon$. 
Let $B$ be such that 
$\bP[\|x - \mu\| \geq B] \leq \epsilon/2$. Also let 
$M$ be the covering number of the unit ball in the dual norm 
$\|\cdot\|_*$. % to $\|\cdot\|$.

Then, given 
$n$ samples $x_1, \ldots, x_n \sim p$, with probability 
$1 - \delta - \exp(-\epsilon n / 6)$ there is a subset $T$ of 
$(1-\epsilon)n$ of the $x_i$ such that $T$ is 
$(\sigma', \epsilon)$-resilient with 
$\sigma' = \oo\Big(\sigma \cdot \Big(1 + \sqrt{\frac{\log(M/\delta)}{\epsilon^2 n}} + \frac{(B/\sigma) \log(M/\delta)}{n}\Big)\Big)$.
\end{proposition}
Note that Proposition~\ref{prop:concentration-resilient} only guarantees resilience 
on a $(1-\epsilon)n$-element subset of the $x_i$, rather than all of $x_1, \ldots, x_n$. 
From the perspective of robust estimation, this is sufficient, as we can simply regard 
the remaining $\epsilon n$ points as part of the ``bad'' points controlled by Alice. 
This weaker requirement seems to be actually necessary to achieve 
Proposition~\ref{prop:concentration-resilient}, and was also exploited in 
\citet{charikar2017learning} to yield improved bounds for a graph partitioning problem. 
There has been a great deal of recent interest in showing how to ``prune'' samples to 
achieve faster rates in random matrix settings 
\citep{guedon2014community,le2015concentration,rebrova2015coverings,rebrova2016norms}, 
and we think the general investigation of such pruning results is likely to be fruitful.

We remark that the sample complexity in Proposition~\ref{prop:concentration-resilient} 
is suboptimal in many cases, requiring roughly $d^{1.5}$ samples when $d$ samples would suffice. 
At the end of the next subsection we discuss a tighter but more specialized bound based on 
spectral graph sparsification.

%This is a concentration result showing that, if the $2k$th moments of a distribution is bounded, 
%then a sample of size roughly $n = \log M$ will have bounded empirical $k$th moments with 
%high probability, after pruning $\epsilon n$ ``bad'' points. Importantly, a set with 
%empirical $k$th moments bounded by $\sigma$ in a norm $\|\cdot\|$ will be 
%$(\oo(\sigma \epsilon^{1-1/k}), \epsilon)$-resilient (in the dual norm $\|\cdot\|_*$) 
%whenever $k \geq 1$. \todo{check $k < 1$ case} \todo{remark $k$ need not be integral}

\paragraph{Applications.} Propositions~\ref{prop:resilience} and 
\ref{prop:concentration-resilient} 
together give us a powerful tool for deriving information-theoretic robust recovery results: 
one needs simply establish resilience for the population distribution $p$, 
then use Proposition~\ref{prop:concentration-resilient} to obtain finite sample bounds and 
Proposition~\ref{prop:resilience} to obtain robust recovery guarantees.
We do this in three illustrative settings: $\ell_2$ mean estimation, 
learning discrete distributions, and stochastic block models.
We outline the results below; formal statements and proofs are deferred to 
Section~\ref{sec:info-theory-applications}.

\textbf{Mean estimation in $\ell_2$-norm.}
Suppose that a distribution on $\bR^d$ has 
bounded $k$th moments: 
$\bE_{x \sim p}[|\langle x-\mu, v \rangle|^{k}]^{1/k} \leq \sigma \|v\|_2$ for all $v$ for some 
$k \geq 2$. Then $p$ is $(\oo(\sigma \epsilon^{1-1/k}), \epsilon)$-resilient in 
the $\ell_2$-norm. 
Propositions~\ref{prop:concentration-resilient} and \ref{prop:resilience} then imply that, 
given $n \geq \frac{d^{1.5}}{\epsilon} + \frac{d}{\epsilon^{2}}$ samples from $p$, 
and an $\epsilon$-fraction of corruptions, it is possible to recover the mean to $\ell_2$-error 
$\oo(\sigma\epsilon^{1-1/k})$. Moreover, if only an $\alpha$-fraction of points 
are good, the mean can be recovered to error 
$\oo(\sigma\alpha^{-1/k})$ with probability $\Omega(\alpha)$.

The $d^{1.5}/\epsilon$ term in the sample complexity is likely loose, 
and we believe the true dependence on $d$ is at most $d\log(d)$. This looseness comes from 
Proposition~\ref{prop:concentration-resilient}, which uses a na\"ive covering argument 
and could potentially be improved with more sophisticated tools. Nevertheless, it is interesting 
that resilience holds long before the empirical $k$th moments concentrate, which would 
require $d^{k/2}$ samples.

\textbf{Distribution learning.}
Suppose that we are given $k$-tuples of independent 
samples from a discrete distribution: $p = \pi^k$, where $\pi$ is a distribution 
on $\{1,\ldots,m\}$. By taking the empirical average of the $k$ samples from $\pi$, 
we can treat a sample from $p$ as an element in the $m$-dimensional simplex 
$\Delta_m$. This distribution turns out to be resilient in the $\ell_1$-norm with 
$\sigma = \oo(\epsilon \sqrt{\log(1/\epsilon)/k})$, which allows us to estimate $p$ 
in the $\ell_1$-norm (i.e., total variation norm) and recover $\hat{\pi}$ such that 
$\|\hat{\pi} - \pi\|_{\TV} = \oo(\epsilon\sqrt{\log(1/\epsilon)/k})$. 
This reveals a pleasing ``error correction'' property: if we are given $k$ samples at a 
time, either all or none of which are good, then our error is $\sqrt{k}$ times smaller than if 
we only observe the samples individually.

\textbf{Stochastic block models.}
Finally, we consider the \emph{semi-random stochastic block model} studied in 
\citet{charikar2017learning} (described in detail in Section~\ref{sec:sbm-app}). 
For a graph on $n$ vertices, this model posits a subset $S$ of $\alpha n$ ``good'' vertices, 
which are connected to each other with probability $\frac{a}{n}$ and to the other 
(``bad'') vertices with probability $\frac{b}{n}$ (where $b < a$); the connections among the bad vertices 
can be arbitrary. The goal is to recover the set $S$.

We think of each row of the adjacency matrix as a vector in $\{0,1\}^n$, 
and show that for the good vertices these vectors are resilient in a truncated 
$\ell_1$-norm $\|x\|$, defined as the sum of the $\alpha n$ largest coordinates 
of $x$ (in absolute value). 
In this case, we have $\sigma = \oo(\alpha \sqrt{a\log(2/\alpha)})$ (this requires 
a separate argument from Proposition~\ref{prop:concentration-resilient} to get tight bounds).
Applying Proposition \ref{prop:resilience}, we find that we are able to 
recover (with probability $\frac{\alpha}{2}$) a set $\hat{S}$ with 
\begin{equation}
\frac{1}{\alpha n}|S \triangle \hat{S}| = \oo\p{\frac{a\log(2/\alpha)}{(a-b)^2\alpha^2}}.
\end{equation} 
In particular, we get non-trivial guarantees as long as 
$\frac{(a-b)^2}{a} \gg \frac{\log(2/\alpha)}{\alpha^2}$. 
\citet{charikar2017learning} derive a weaker (but computationally efficient) bound when 
$\frac{(a-b)^2}{a} \gg \frac{\log(2/\alpha)}{\alpha^3}$, and remark on the 
similarity to the famous \emph{Kesten-Stigum threshold} 
$\frac{(a-b)^2}{a} \gg \frac{1}{\alpha^2}$, which is the conjectured threshold 
for computationally efficient recovery in the classical stochastic block model (see 
\citet{decelle2011asymptotic} for the conjecture, and \citet{mossel2013proof,massoulie2014community} for a proof in the two-block case).
Our information-theoretic 
upper bound matches the Kesten-Stigum threshold up to a $\log(2/\alpha)$ factor. 
We conjecture that this upper bound is tight; 
%, i.e. that the Kesten-Stigum threshold is the correct \emph{information-theoretic} 
%threshold in the semi-random setting; 
some evidence for this is given in 
\citet{steinhardt2017clique}, which provides a nearly matching information-theoretic 
lower bound when $a = 1$, $b = \frac{1}{2}$.

\subsection{Strong Convexity, Second Moments, and Efficient Algorithms}
\label{sec:algo-intro}

Most existing algorithmic results on robust mean estimation rely on analyzing 
the empirical covariance of the data in some way (see, e.g., 
\citet{lai2016agnostic,diakonikolas2016robust,balakrishnan2017sparse}).
In this section we establish connections between bounded covariance and resilience, 
and show that in a very general sense, bounded covariance is indeed sufficient to enable 
robust mean estimation.

Given a norm $\|\cdot\|$, we say that a set of points $x_1, \ldots, x_n$ has 
\emph{variance bounded by $\sigma_0^2$} in that norm if 
$\frac{1}{n} \sum_{i=1}^n \langle x_i - \mu, v \rangle^2 \leq \sigma_0^2 \|v\|_*^2$ 
(recall $\|\cdot\|_*$ denotes the dual norm).
Since this implies a tail bound along every direction, it is easy to see 
(c.f.~Lemma~\ref{lem:tail-bound}) that a set with variance bounded by $\sigma_0^2$ 
is $(\oo(\sigma_0\sqrt{\epsilon}), \epsilon)$-resilient around its mean for all $\epsilon < \frac{1}{2}$.
Therefore, bounded variance implies resilience.

An important result is that the converse is also true, 
\emph{provided the norm is strongly convex}. We say that a norm $\|\cdot\|$ is 
$\gamma$-strongly convex if $\|x+y\|^2 + \|x-y\|^2 \geq 2(\|x\|^2 + \gamma \|y\|^2)$ for 
all $x, y \in \bR^d$.\footnote{In the language of Banach space theory, this is also referred 
to as having {bounded co-type}.}
As an example, the $\ell_p$-norm is $({p-1})$-strongly convex 
for $p \in (1,2]$. For strongly convex norms, we show that any resilient set has a highly 
resilient ``core'' with bounded variance: %$2$nd moments:
\begin{theorem}
\label{thm:powering-up-intro}
If $S$ is $(\sigma,\frac{1}{2})$-resilient in a $\gamma$-strongly 
convex norm $\|\cdot\|$, then $S$ contains a set $S_0$ of size 
at least $\frac{1}{2}|S|$ with bounded variance:
$\frac{1}{|S_0|} \sum_{i \in S_0} \langle x_i - \mu, v \rangle^2 \leq \frac{288\sigma^2}{\gamma}\|v\|_*^2$ for all $v$.
\end{theorem}
Using Lemma~\ref{lem:tail-bound}, we can show that 
$(\sigma,\frac{1}{2})$-resilience is equivalent to having bounded $1$st moments 
in every direction; Theorem~\ref{thm:powering-up-intro} can thus be interpreted as saying that 
any set with bounded $1$st moments can be pruned to have bounded $2$nd moments.

We found this result quite striking---the fact that Theorem~\ref{thm:powering-up-intro} 
can hold with no dimension-dependent factors is far from obvious.
In fact, if we replace 2nd moments with 3rd moments or take a non-strongly-convex norm 
then the analog of Theorem~\ref{thm:powering-up-intro} is false: we incur 
polynomial factors in the dimension even if $S$ is the standard basis of $\bR^d$
(see Section~\ref{sec:counterexample} for details).
The proof of Theorem~\ref{thm:powering-up-intro} involves 
minimax duality and Khintchine's inequality.
We can also strengthen Theorem~\ref{thm:powering-up-intro} to yield $S_0$ 
of size $(1-\epsilon)|S|$.
The proofs of both results %Theorem~\ref{thm:powering-up-intro} 
are given in Section~\ref{sec:powering-up} 
and may be of independent interest.
%\todo{reference proofs}

\paragraph{Algorithmic results.}
Given points with bounded variance, 
we establish algorithmic 
results assuming that one can solve the ``generalized eigenvalue'' problem 
$\max_{\|v\|_* \leq 1} v^{\top}Av$ up to some multiplicative accuracy $\kappa$. 
Specifically, we make the following assumption:
\begin{assumption}[$\kappa$-Approximability]
\label{ass:opt}
There is a convex set $\sP$ of PSD matrices such that 
\begin{equation}
\sup_{\|v\|_* \leq 1} v^{\top}Av \leq \sup_{M \in \sP} \langle A, M \rangle \leq \kappa \sup_{\|v\|_* \leq 1} v^{\top}Av
\end{equation}
for every PSD matrix $A$. 
Moreover, it is possible to optimize linear functions over $\sP$ in polynomial time.
\end{assumption}
A result of \citet{nesterov1998semidefinite} implies that this is true with 
$\kappa = \oo(1)$ if $\|\cdot\|_*$ is any ``quadratically convex'' norm, 
which includes the $\ell_q$-norms for $q \in [2, \infty]$.
Also, while we do not use it in this paper, one can sometimes exploit weaker versions of 
Assumption~\ref{ass:opt} that only require $\sup_{M \in \sP} \langle A, M \rangle$ to be 
small for certain matrices $A$; see for instance \citet{li2017sparse}, which obtains an 
algorithm for robust sparse mean estimation even though Assumption~\ref{ass:opt} (as well as 
strong convexity) fails to hold in that setting.

Our main algorithmic result is the following:
\begin{theorem}
\label{thm:alg-intro}
Suppose that $x_1, \ldots, x_n$ contains a subset $S$ of size $(1-\epsilon)n$ whose variance around 
its mean $\mu$ is bounded by $\sigma_0^2$ in the norm $\|\cdot\|$. 
Also suppose that Assumption~\ref{ass:opt} holds for the dual norm $\|\cdot\|_*$. 
Then, if $\epsilon \leq \frac{1}{4}$, there is a polynomial-time algorithm whose output satisfies 
$\|\hat{\mu} - \mu\| = \oo\big(\sigma_0 \sqrt{\kappa \epsilon}\big)$. 

If, in addition, $\|\cdot\|$ is $\gamma$-strongly convex, then even if $S$ 
only has size $\alpha n$ there is a polynomial-time algorithm such that 
$\|\hat{\mu} - \mu\| = \oo\big(\frac{\sqrt{\kappa} \sigma_0}{\sqrt{\gamma} \alpha}\big)$ 
with probability $\Omega(\alpha)$.
\end{theorem}
This is essentially a more restrictive, but computationally efficient version of 
Proposition~\ref{prop:resilience}. We note that for the $\ell_2$-norm, the algorithm can be 
implemented as an SVD (singular value decomposition) combined with a filtering step; for 
more general norms, the SVD is replaced with a semidefinite program.

In the small-$\epsilon$ regime, Theorem~\ref{thm:alg-intro} is in line with 
existing results which typically achieve errors of $\oo(\sqrt{\epsilon})$ in specific norms 
(see for instance a concurrent result of \citet{diakonikolas2017practical}). 
While 
several papers achieve stronger rates of $\oo(\epsilon^{3/4})$ \citep{lai2016agnostic} or 
$\tilde{\oo}(\epsilon)$ \citep{diakonikolas2016robust,balakrishnan2017sparse}, 
these stronger results rely crucially on specific distributional 
assumptions such as Gaussianity. At the time of writing of this paper, no 
results obtained rates better than $\oo(\sqrt{\epsilon})$ for any general class of distributions 
(even under strong assumptions 
such as sub-Gaussianity). After initial publication of this paper, \citet{kothari2017agnostic} 
surpassed $\sqrt{\epsilon}$ and obtained rates of $\epsilon^{1-\gamma}$ for any 
$\gamma > 0$, for distributions satisfying the Poincar\'{e} isoperimetric inequality.

In the small-$\alpha$ regime, Theorem~\ref{thm:alg-intro} generalizes the mean estimation 
results of \citet{charikar2017learning} to norms beyond the $\ell_2$-norm. That paper achieves 
a better rate of $1/\sqrt{\alpha}$ (versus the $1/\alpha$ rate given here); 
we think it is likely possible to achieve the $1/\sqrt{\alpha}$ rate in general, but 
leave this for future work.

\paragraph{Applications.} 
Because Assumption~\ref{ass:opt} 
holds for $\ell_p$-norms, we can perform robust estimation in $\ell_p$-norms for any 
$p \in [1,2]$, as long as the data have bounded variance 
in the dual $\ell_q$-norm (where $\frac{1}{p} + \frac{1}{q} = 1$); see 
Corollary~\ref{cor:lp} for a formal statement. 
This is the first efficient algorithm for performing robust mean estimation in any 
$\ell_p$-norm with $p \neq 2$. The $\ell_1$-norm in particular is often a more meaningful 
metric than the $\ell_2$-norm in discrete settings, allowing us to improve on existing 
results. 
%In particular, it yields new results for robustly 
%learning discrete distributions (see Corollary~\ref{cor:dist-learning-alg}).

Indeed, as in the previous section, suppose we 
are given $k$-tuples of samples from a discrete distribution $\pi$ on $\{1,\ldots,m\}$. 
Applying Theorem~\ref{thm:alg-intro} with the $\ell_1$-norm yields an algorithm 
recovering a $\hat{\pi}$ with $\|\hat{\pi} - \pi\|_{TV} = \tilde{\oo}(\sqrt{\epsilon/k})$.\footnote{The $\tilde{\oo}$ notation suppresses log factors in $m$ and $\epsilon$; the dependence on $m$ can likely be removed with a more careful analysis.}
% but we present this version of the result for simplicity.} 
In contrast, bounds using the $\ell_2$-norm would only yield 
$\|\hat{\pi} - \pi\|_2 = \oo(\sqrt{\epsilon \pi_{\max}/k})$, which is 
substantially weaker when the maximum probability $\pi_{\max}$ is large.
Our result has a similar flavor to that of \citet{diakonikolas2016robust} on robustly 
estimating binary product distributions, for which directly applying 
$\ell_2$ mean estimation was also insufficient.
We discuss our bounds in more detail in Section~\ref{sec:algo-applications}.

\textbf{Finite-sample bound.} To get the best sample complexity for the applications 
above, we provide an additional finite-sample bound focused on showing that a set of 
points has bounded variance. This is a simple but useful generalization of 
Proposition B.1 of \citet{charikar2017learning}; it shows that in a very generic sense, 
given $d$ samples from a distribution on $\bR^d$ with bounded population variance, we can find a 
subset of samples with bounded variance with high probability. It involves 
pruning the samples in a non-trivial way based on ideas from graph sparsification 
\citep{batson2012twice}. The formal statement is given in Section~\ref{sec:concentration-cov}.

\subsection{Low-Rank Recovery}
\label{sec:rank-intro}

Finally, to illustrate that the idea behind resilience is quite general and not restricted to 
mean estimation, we also provide results on recovering a rank-$k$ approximation 
to the data in the presence of arbitrary outliers. %We first define the goal:
Given a set of points $[x_i]_{i \in S}$, let $X_S$ be the matrix whose 
columns are the $x_i$. 
Our goal is to obtain a low-rank matrix $P$ such that the operator norm 
$\|(I-P)X_S\|_2$ is not much larger 
than $\sigma_{k+1}(X_S)$, where $\sigma_{k+1}$ denotes the $k+1$st singular value; we wish 
to do this even if $S$ is corrupted to a set $\tilde{S}$ by adding arbitrary outliers.

As before, we start by formulating an appropriate resilience criterion: 
\begin{definition}[Rank-resilience]
\label{def:resilience-rank-k}
A set of points $[x_i]_{i \in S}$ in $\bR^d$ 
is \emph{$\delta$-rank-resilient} if
%the matrix 
%$X = [x_i]_{i \in S}$ satisfies (i) $\sigma_{k+1}(X) \leq \sigma\sqrt{|S|}$
%and (ii) 
for all subsets $T$ of size at least 
$(1-\delta)|S|$, we have $\col(X_T) = \col(X_S)$ 
and $\|X_T^{\dagger}X_S\|_{2} \leq 2$, where $\dagger$ is the pseudoinverse and 
$\col$ denotes column space.
\end{definition}
Rank-resilience says that the variation in $X$ 
should be sufficiently spread out: there should not be a direction 
of variation that is concentrated in only a $\delta$-fraction of 
the points.
Under rank-resilience, we can perform efficient rank-$k$ recovery even 
in the presence of a $\delta$-fraction of arbitrary data:
\begin{theorem}
\label{thm:main-rank-k}
Let $\delta\leq \frac{1}{3}$. 
If a set of $n$ points contains a set $S$ of size 
$(1-\delta)n$ that is $\delta$-rank-resilient, 
then it is possible to efficiently recover 
%there is an efficient algorithm recovering 
a matrix $P$ of rank at most $15k$
such that $\|(I-P)X_S\|_{2} = \oo(\sigma_{k+1}(X_S))$.
\end{theorem}
The power of Theorem~\ref{thm:main-rank-k} comes from the fact that 
the error depends on $\sigma_{k+1}$ rather than e.g.~$\sigma_2$, which 
is what previous results yielded.
This distinction is crucial in practice, since most data have a few 
(but more than one) large singular values followed by many small singular values. %that are much larger than the remainder.
Note that in contrast to Theorem~\ref{thm:alg-intro}, Theorem~\ref{thm:main-rank-k} 
only holds when $S$ is relatively large: at least $(1-\delta)n \geq \frac{2}{3}n$ in size.

\subsection{Summary, Technical Highlights, and Roadmap}
\label{sec:roadmap}

In summary, we have provided a deterministic condition on a set of points 
that enables robust mean estimation, and provided finite-sample bounds showing 
that this condition holds with high probability in many concrete settings. 
This yields new results for distribution learning, stochastic block models, 
mean estimation under bounded moments, and mean estimation in $\ell_p$ norms.
We also provided an extension of our condition that yields results for robust low-rank recovery.

Beyond the results themselves, the following technical aspects of our work may 
be particularly interesting: The proof of Proposition~\ref{prop:resilience} (establishing 
that resilience is indeed sufficient for robust estimation), while simple, is a nice 
pigeonhole argument that we found to be conceptually illuminating. 

In addition, the proof of Theorem~\ref{thm:powering-up-intro}, on pruning resilient sets to 
obtain sets with bounded variance, exploits strong convexity in a non-trivial way in conjunction 
with minimax duality; we think it reveals a fairly non-obvious geometric structure 
in resilient sets, and also shows how the ability to prune points can yield sets with 
meaningfully stronger properties.

Finally, in the proof of our algorithmic result (Theorem~\ref{thm:alg-intro}), we establish
an interesting generalization of the inequality 
$\sum_{i,j} X_{ij}^2 \leq \rank(X) \cdot \|X\|_2^2$, which holds not just 
for the $\ell_2$-norm but for any strongly convex norm. This is given as 
Lemma~\ref{lem:op-fro-p}.
% definition of resilience
% highly stable core
% concentration result -- general result, and BSS argument

\paragraph{Roadmap.} The rest of the paper is organized as follows. 
In Section~\ref{sec:info-theory}, we prove our information-theoretic recovery result 
for resilient sets (Proposition~\ref{prop:resilience}). % presented in Section~\ref{sec:info-theory-intro}. 
In Section~\ref{sec:powering-up}, 
we prove Theorem~\ref{thm:powering-up-intro} establishing that all resilient sets 
in strongly convex norms contain large subsets with bounded variance; we also prove 
a more precise version of Theorem~\ref{thm:powering-up-intro} in 
Section~\ref{sec:powering-up-2}. In Section~\ref{sec:alg}, we prove our algorithmic 
results, warming up with the $\ell_2$-norm (Section~\ref{sec:main-l2}) 
and then moving to general norms (Section~\ref{sec:main-lp}). 
In Section~\ref{sec:rank-k}, we prove our results on rank-$k$ recovery. 
In Section~\ref{sec:concentration}, we present and prove the finite-sample bounds 
discussed in Section~\ref{sec:intro-concentration}. 
In Sections~\ref{sec:info-theory-applications} and \ref{sec:algo-applications}, 
we provide applications of our information-theoretic and algorithmic results, 
respectively.

\input related-work

%% file: related-work.tex
\subsection{Related Work}
\label{sec:related-work}
A number of authors have recently studied robust estimation and learning in high-dimensional 
settings: \citet{lai2016agnostic} study mean and covariance estimation, while 
\citet{diakonikolas2016robust} focus on estimating Gaussian and binary product distributions, 
as well as mixtures thereof; note that this implies mean/covariance estimation of the 
corresponding distributions.
\citet{charikar2017learning} recently showed that robust 
estimation is possible even when the fraction $\alpha$ of ``good'' data is less than 
$\frac{1}{2}$. We refer to these papers for an overview of the broader robust estimation 
literature; since those papers, a number of additional results have also been published:
%but highlight in particular \citet{bhatia2015robust}, who study 
%linear regression, as well as \citet{steinhardt2016avoiding}, who study a crowdsourcing setting. 
%In addition, 
\citet{diakonikolas2017practical} provide a case study of various robust 
estimation methods in a genomic setting, 
\citet{balakrishnan2017sparse} study sparse mean estimation, and others have studied problems 
including regression, Bayes nets, planted clique, and several other settings 
\citep{diakonikolas2016bayes,diakonikolas2016statistical,diakonikolas2017robustly,diakonikolas2017learning,kane2017robust,meister2017data}.

Special cases of the resilience criterion are implicit in some of these earlier works; 
for instance, $\ell_2$-resilience appears in equation (9) in 
\citet{diakonikolas2016robust}, and resilience in a sparsity-inducing norm appears 
in Theorem 4.5 of \citet{li2017sparse}. 
%which was published shortly before an initial version of this paper. 
However, these conditions typically appear concurrently with other stronger 
conditions, and the general sufficiency of resilience for information-theoretic recovery 
appears to be unappreciated (for instance, \citet{li2017sparse}, despite already 
having implicitly established resilience, proves its information-theoretic 
results via reduction to a tournament lemma from \citet{diakonikolas2016robust}).

%We can compare our results to what is known for estimation in the $\ell_2$-norm.
%Here, when $\alpha \approx 1$, \citet{lai2016agnostic} obtain error 
%$\oot\p{\sigma\sqrt{1-\alpha}}$ assuming bounded $4$th moments, while 
%\citet{diakonikolas2016robust} obtain error $\oot\p{\sigma\cdot(1-\alpha)}$ under 
%sub-Gaussianity. % assumptions. 
%Our criterion for obtaining error 
%$\oot\p{\sigma\sqrt{1-\alpha}}$ is weaker than bounded $2$nd moments, 
%and so improves upon \citet{lai2016agnostic}.
%
%When $\alpha < \frac{1}{2}$, the only point of comparison is 
%\citet{charikar2017learning}, who obtain error
%$\oo\p{\frac{\sigma}{\sqrt{\alpha}}}$ under bounded $1$st moments. 
%We obtain a bound of $\oo\p{\frac{\sigma}{\alpha}}$ under $(\sigma,\frac{1}{2})$-resilience, 
%which essentially corresponds to bounded $1$st moments as well. 
%Our results are thus weaker in the $\ell_2$-norm but hold more generally.
%We think \citeauthor{charikar2017learning}'s argument could be extended to $\ell_p$ norms, 
%but only with considerable 
%effort. Our approach is simple enough that the generalization 
%to $\ell_p$ norms is obvious.

Low rank estimation was studied by \citet{lai2016agnostic}, but their 
bounds depend on the maximum eigenvalue $\|\Sigma\|_2$ of the covariance matrix, 
while our bound provides robust recovery guarantees in terms 
of lower singular values of $\Sigma$. (Some work, such as \citet{diakonikolas2016robust}, 
shows how to estimate all of $\Sigma$ in e.g. Frobenius norm, but appears to require 
the samples to be drawn from a Gaussian.)

%% file: info-theory.tex
\section{Resilience and Robustness: Information-Theoretic Sufficiency}
\label{sec:info-theory}
\label{sec:resilience-proof}

Recall the definition of resilience: $S$ is 
$(\sigma,\epsilon)$-resilient if 
$\|\frac{1}{|T|} \sum_{i \in T} (x_i - \mu)\| \leq \sigma$ 
whenever $T \subseteq S$ and $|T| \geq (1-\epsilon)|S|$. 
Here we establish Proposition~\ref{prop:resilience} showing that, 
if we ignore computational efficiency, resilience 
leads directly to an algorithm for robust mean estimation. 
\begin{prooff}{Proposition~\ref{prop:resilience}}
We prove Proposition~\ref{prop:resilience} via a constructive (albeit exponential-time) 
algorithm. To prove the first part, suppose that the true set $S$ is 
$(\sigma,\frac{\epsilon}{1-\epsilon})$-resilient around $\mu$, and let $S'$ be any 
set of size $(1-\epsilon)n$ that is $(\sigma,\frac{\epsilon}{1-\epsilon})$-resilient 
(around some potentially different vector $\mu'$). 
We claim that $\mu'$ is sufficiently close to $\mu$.

Indeed, let $T = S \cap S'$, which by the pigeonhole principle has 
size at least $(1-2\epsilon)n = \frac{1-2\epsilon}{1-\epsilon}|S| = (1-\frac{\epsilon}{1-\epsilon})|S|$. Therefore, by the definition of resilience,
\begin{equation}
\textstyle\left\|\frac{1}{|T|} \sum_{i \in T} (x_i - \mu)\right\| \leq \sigma.
%\sigmabest(\frac{1-\alpha}{\alpha}).
\end{equation}
But by the same argument, $\|\frac{1}{|T|} \sum_{i \in T} (x_i - \mu') \| \leq \sigma$ as well.
%best(\frac{1-\alpha}{\alpha})$ as well. 
By the triangle inequality, $\|\mu - \mu'\| \leq 2\sigma$, %best(\frac{1-\alpha}{\alpha})$, 
which completes the first part of the proposition.

For the second part, we need the following simple lemma relating 
$\epsilon$-resilience to $(1-\epsilon)$-resilience:
\begin{lemma}
\label{lem:reverse}
For any $0 < \epsilon < 1$, a distribution/set is $(\sigma, \epsilon)$-resilient around 
its mean $\mu$ if and only if it is $(\frac{1-\epsilon}{\epsilon}\sigma, 1-\epsilon)$-resilient. 
More generally, even if $\mu$ is not the mean then the distribution/set is 
$(\frac{2-\epsilon}{\epsilon}\sigma, 1-\epsilon)$-resilient.
%then it is $(\frac{2}{\epsilon}\sigma, 1-\epsilon)$-resilient. 
In other words, if $\|\frac{1}{|T|} \sum_{i \in T} (x_i - \mu)\| \leq \sigma$ 
for all sets $T$ of size at least $(1-\epsilon)n$, then 
$\|\frac{1}{|T'|} \sum_{i \in T'} (x_i - \mu)\| \leq \frac{2-\epsilon}{\epsilon} \sigma$ 
for all sets $T'$ of size at least $\epsilon n$.
\end{lemma}

Given Lemma~\ref{lem:reverse}, 
the second part of Proposition~\ref{prop:resilience} is similar to the first part, 
but requires us to consider multiple resilient sets $S_i$ rather than a single $S'$.
Suppose $S$ is $(\sigma,\frac{\alpha}{4})$-resilient around $\mu$--and 
thus also $(\frac{8}{\alpha}\sigma, 1-\frac{\alpha}{4})$-resilient by Lemma~\ref{lem:reverse}--and 
let $S_1, \ldots, S_m$ be a maximal collection of subsets of $[n]$ such that:
\begin{enumerate}
\setlength\itemsep{0em}
\item $|S_j| \geq \frac{\alpha}{2} n$ for all $j$.
\item $S_j$ is $(\frac{8}{\alpha}\sigma,1-\frac{\alpha}{2})$-resilient 
around some point $\mu_j$.
%\item $|S_j \cap S_{j'}| \leq \frac{\alpha^2}{2}n$ for all $j \neq j'$.
\item $S_j \cap S_{j'} = \emptyset$ for all $j \neq j'$.
\end{enumerate}
%Let $\mu_i$ be the mean vector corresponding to $S_i$. 
Clearly $m \leq \frac{2}{\alpha}$. 
We claim that at least 
one of the $\mu_j$ is close to $\mu$. By maximality of the collection $\{S_j\}_{j=1}^m$, 
it must be that $S_0 = S \backslash (S_1 \cup \cdots \cup S_m)$ cannot be added to the collection.
First suppose that $|S_0| \geq \frac{\alpha}{2}n$. Then $S_0$ is 
$(\frac{8}{\alpha} \sigma, 1-\frac{\alpha}{2})$-resilient (because any subset of 
$\frac{\alpha}{2} |S_0|$ points in $S_0$ is a subset of at least $\frac{\alpha}{4}|S|$ points in $S$).
But this contradicts the maximality of $\{S_j\}_{j=1}^m$, so we must have 
$|S_0| < \frac{\alpha}{2}n$.

Now, this implies that $|S \cap (S_1 \cup \cdots \cup S_m)| \geq \frac{\alpha}{2}n$, so by 
pigeonhole we must have $|S \cap S_j| \geq \frac{\alpha}{2}|S_j|$ for some $j$. Letting 
$T = S \cap S_j$ as before, we find that $|T| \geq \frac{\alpha}{2}|S_j| \geq \frac{\alpha}{4}|S|$ and hence by resilience of $S_j$ and $S$ we have
$\|\mu - \mu_j\| \leq 2 \cdot (\frac{8}{\alpha} \sigma) = \frac{16}{\alpha} \sigma$. 
%by Lemma~\ref{lem:reverse}. %\sigmabest(1-\frac{\alpha}{2})$.
If we output one of the $\mu_j$ at random, 
we are then within the desired distance of $\mu$ with probability $\frac{1}{m} \geq \frac{\alpha}{2}$.
%
%It remains to bound $m$. By the principle of inclusion-exclusion, we have 
%$n \geq |S_1 \cup \cdots \cup S_m| \geq \sum_{j=1}^m |S_j| - \sum_{1 \leq j < j' \leq m} |S_j \cap S_{j'}| \geq \alpha nm - \frac{\alpha^2n}{2}\binom{m}{2}$. 
%Simple algebra shows that $m \leq \frac{2}{\alpha}$, which completes the proof.
\end{prooff}

%% file: powering-up.tex
%\vskip -0.4in
%\phantom{x}

\section{Powering up Resilience: Finding a Core with Bounded Variance}
\label{sec:powering-up}

In this section we prove 
%the key ``powering up'' result 
Theorem~\ref{thm:powering-up-intro}, which says that for strongly convex norms, 
every resilient set contains a core with bounded variance.
Recall that this is important for enabling algorithmic applications that depend on 
a bounded variance condition.
%While the previous section focused on the $\ell_2$ case, this section 
%and the next will focus on general $\ell_p$ norms ($1 < p \leq 2$).

First recall the definition of resilience (Definition~\ref{def:resilience-lp}):
a set $S$ is $(\sigma,\epsilon)$-resilient if for every set 
$T \subseteq S$ of size $(1-\epsilon)|S|$, we have 
$\|\frac{1}{|T|} \sum_{i \in T} (x_i - \mu)\| \leq \sigma$. For 
$\epsilon = \frac{1}{2}$, we observe that resilience in a norm is equivalent to 
having bounded first moments in the dual norm:
\begin{lemma}
\label{lem:1st-moment-lp}
Suppose that $S$ is $(\sigma,\frac{1}{2})$-resilient in a norm $\|\cdot\|$, and 
let $\|\cdot\|_*$ be the dual norm. 
Then $S$ has 1st moments bounded by $3\sigma$: 
$\frac{1}{|S|} \sum_{i \in S} |\langle x_i - \mu, v \rangle| \leq 3\sigma \|v\|_*$ 
for all $v \in \bR^d$.

Conversely, if $S$ has 1st moments bounded by $\sigma$, 
it is $(2\sigma,\frac{1}{2})$-resilient.
\end{lemma}

The proof is routine and can be found in Section~\ref{sec:1st-moment-lp-proof}.
Supposing a set has bounded $1$st moments, we will show that it has a large core with 
bounded second moments. %Unlike Lemma~\ref{lem:1st-moment-lp}, 
This next result is \emph{not} routine:
\begin{proposition}
\label{prop:1st-2nd-lp}
\label{prop:powering-up-lp}
Let $S$ be any set with $1$st moments bounded by $\sigma$.
Then if the norm $\|\cdot\|$ is $\gamma$-strongly convex, 
there exists a core $S_0$ of size at least 
$\frac{1}{2}|S|$ with variance bounded by $\frac{32\sigma^2}{\gamma}$. %,e^2\log(d))$. 
That is, $\frac{1}{|S_0|} \sum_{i \in S_0} |\langle x_i - \mu, v \rangle|^2 \leq 
\frac{32\sigma^2}{\gamma} \|v\|_*^2$ \,%,e^2\log(d))\|v\|_q^2$ 
for all $v \in \bR^d$.
\end{proposition}
The assumptions seem necessary:
i.e., such a core does not exist when $\|\cdot\|$ is the $\ell_p$-norm with 
$p > 2$ (which is a non-strongly-convex norm),
or with bounded $3$rd moments for $p = 2$ (see Section~\ref{sec:counterexample}).
The proof of Proposition~\ref{prop:1st-2nd-lp} uses minimax duality 
and Khintchine's inequality \citep{khintchine1923uber}. Note that 
Lemma~\ref{lem:1st-moment-lp} and Proposition~\ref{prop:1st-2nd-lp} together imply 
Theorem~\ref{thm:powering-up-intro}.

\begin{prooff}{Proposition~\ref{prop:1st-2nd-lp}}
Without loss of generality take $\mu = 0$ and suppose that $S = [n]$.
We can pose the problem of finding a resilient core as an integer program:
\begin{equation}
\label{eq:minimax-0}
\min_{c \in \{0,1\}^n, \|c\|_1 \geq \frac{n}{2}} \max_{\|v\|_* \leq 1} \frac{1}{n} \sum_{i=1}^n c_i |\langle x_i, v \rangle|^2.
\end{equation}
Here the variable $c_i$ indicates whether the point $i$ lies in the core $S_0$.
By taking a continuous relaxation and applying a standard duality argument, we obtain 
the following:
\begin{lemma}
\label{lem:minimax}
%Let $\Delta_m$ be the $m$-dimensional probability simplex. 
Suppose that for all $m$ and all 
vectors $v_1, \ldots, v_m$ satisfying $\sum_{j=1}^m \|v_j\|_*^2 \leq 1$, 
we have
%for all $m$, $\alpha \in \Delta_m$, and $v_{1:m} \in \bR^d$ with 
%$\|v_j\|_q \leq 1$, we have 
\begin{equation}
\label{eq:dual}
%\max_{\substack{\alpha \in \Delta_m \\ \|v_j\|_q \leq 1}} 
\frac{1}{n} \sum_{i=1}^n \sqrt{\sum_{j=1}^m |\langle x_i, v_j \rangle|^2} \leq B.
\end{equation}
%for all weight vectors $\alpha \in \Delta_m$ and vectors $v_{1:m}$ satisfying 
%$\|v_j\|_q \leq 1$.
Then the value of \eqref{eq:minimax-0} is at most $8B^2$.
\end{lemma}
The proof is straightforward and deferred to Section~\ref{sec:minimax-proof}.
Now, to bound \eqref{eq:dual}, 
let $s_1, \ldots, s_m \in \{-1,+1\}$ be i.i.d.~random sign variables. We have
\begin{align}
%B(\alpha_{1:m}, v_{1:m}) &= 
\frac{1}{n} \sum_{i=1}^n \sqrt{\sum_{j=1}^m |\langle x_i, v_j \rangle|^2} 
 &\stackrel{(i)}{\leq} \bE_{s_{1:m}}\Bigg[\frac{\sqrt{2}}{n} \sum_{i=1}^n \Big|\sum_{j=1}^m s_j \langle x_i, v_j \rangle\Big|\Bigg] \\
 &= \bE_{s_{1:m}}\Bigg[\frac{\sqrt{2}}{n} \sum_{i=1}^n \Big|\Big\langle x_i, \sum_{j=1}^m s_j v_j \Big\rangle\Big|\Bigg] \\
 &\stackrel{(ii)}{\leq} \bE_{s_{1:m}}\Bigg[\sqrt{2}\sigma\Big\|\sum_{j=1}^m s_jv_j\Big\|_*\Bigg] \\
\label{eq:q-norm} &\leq \sqrt{2}\sigma{\bE_{s_{1:m}}\Big[\Big\|\sum_{j=1}^m s_jv_j\Big\|_*^2\Big]^{\frac{1}{2}}}.
\end{align}
Here (i) is Khintchine's inequality \citep{haagerup1981best} and (ii) is the assumed first moment bound.
It remains to bound \eqref{eq:q-norm}.
%The key is the following inequality, which amounts to asserting smoothness of 
%the $\ell_q$-norm, and is well-known (c.f. Lemma 17 of \citet{shalev07online}):
%\begin{lemma}
%\label{lem:cotype}
%For any $x$, $y$, and $q \geq 2$, we have 
%$\frac{1}{2}(\|x+y\|_q^2 + \|x-y\|_q^2) \leq \|x\|_q^2 + (q-1)\|y\|_q^2$.
%\end{lemma}
The key is the following inequality asserting that the dual norm $\|\cdot\|_*$ is 
strongly smooth whenever $\|\cdot\|$ is strongly convex (c.f.~Lemma 17 of \citet{shalev07online}):
\begin{lemma}
\label{lem:cotype}
If $\|\cdot\|$ is $\gamma$-strongly convex, then 
$\|\cdot\|_*$ is $(1/\gamma)$-strongly smooth:
$\frac{1}{2}(\|v+w\|_*^2 + \|v-w\|_*^2) \leq \|v\|_*^2 + (1/\gamma)\|w\|_*^2$.
\end{lemma}
Applying Lemma~\ref{lem:cotype} inductively to 
$\bE_{s_{1:m}}\left[\big\|\sum_{j=1}^m s_jv_j\big\|_*^2\right]$, we obtain 
\begin{align}
\label{eq:cotype-induction}
\bE_{s_{1:m}}\Big[\Big\|\sum_{j=1}^m s_jv_j\Big\|_*^2\Big] 
 &\leq \frac{1}{\gamma} \sum_{j=1}^m \|v_j\|_*^2 
 \leq \frac{1}{\gamma}.
\end{align}
Combining with \eqref{eq:q-norm}, we have the bound 
$B \leq \sigma\sqrt{2/\gamma}$, 
which yields the desired result.
%so that by Lemma~\ref{lem:minimax} 
%the value of \eqref{eq:minimax-0} is bounded by $4B^2 \leq 8(q-1)\sigma^2$, 
%which yields the desired result. %lets us find a core with 2nd moments bounded by $16(q-1)\sigma^2$.
\end{prooff}

\vspace*{-2mm}
\subsection{Finding Resilient Cores when $\alpha \approx 1$}
\label{sec:powering-up-2}

Lemma~\ref{lem:1st-moment-lp} together with Proposition~\ref{prop:1st-2nd-lp}
show that a $(\sigma,\frac{1}{2})$-resilient set has a core with bounded $2$nd moments. 
%%% which combined with Proposition~\ref{prop:recovery-l2} from the previous section 
%%% (as well as its analog for $\ell_p$ norms in the next section) almost yields 
%%% Theorem~\ref{thm:main-lp}.
%%% %our main 
%%% %result (Theorem~\ref{thm:main-lp}) 
%%% %on robust mean estimation in $\ell_p$ norms. 
One piece of looseness is that 
Proposition~\ref{prop:1st-2nd-lp} only exploits resilience for $\epsilon = \frac{1}{2}$, 
and hence is not sensitive to the degree of $(\sigma,\epsilon)$-resilience as $\epsilon \to 0$. 
In particular, it only yields a core $S_0$ of size $\frac{1}{2}|S|$, 
while we might hope to find a much larger core of size $(1-\epsilon)|S|$ for 
some small $\epsilon$.

Here we tighten Proposition~\ref{prop:1st-2nd-lp}
to make use of finer-grained resilience information. Recall that we let $\sigmabest(\epsilon)$ 
denote the resilience over sets of size $(1-\epsilon)|S|$. 
For a given $\epsilon$, our goal is to construct a core $S_0$ of size $(1-\epsilon)|S|$ 
with small second moments. The following key quantity will tell us how small the second 
moments can be:
\begin{equation}
\label{eq:sigma-star}
\sigmastar(\epsilon) \eqdef \sqrt{\int_{\epsilon/2}^{1/2} u^{-2}\sigmabest(u)^2 du}.
\end{equation}
The following proposition, proved in Section~\ref{sec:powering-up-lp-eps-proof}, 
says that $\sigmastar$ controls the 2nd moments of $S_0$:
\begin{proposition}
\label{prop:1st-2nd-lp-eps}
\label{prop:powering-up-lp-eps}
Let $S$ be any resilient set in a $\gamma$-strongly-convex norm.
Then for any $\epsilon \leq \frac{1}{2}$, there exists a core $S_0$ of size
$(1-\epsilon)|S|$ with variance bounded by 
$\oo\p{\sigmastar^2(\epsilon)/\gamma}$. %for $q = \frac{p}{p-1}$.
\end{proposition}
The proof is similar to Proposition~\ref{prop:powering-up-lp}, but requires more careful
bookkeeping.

To interpret $\sigmastar$, suppose that $\sigmabest(\epsilon) = \sigma \epsilon^{1-1/r}$ for some 
$r \in [1,2)$, which roughly corresponds to having bounded $r$th moments. 
Then $\sigmastar^2(\epsilon) = \sigma^2 {\int_{\epsilon/2}^{1/2} u^{-2/r} du} \leq 
\frac{\sigma^2}{{2/r-1}} \p{\frac{2}{\epsilon}}^{2/r-1}$.
If $r = 1$ then a core of size $(1-\epsilon)|S|$ might require second moments as 
large as $\frac{\sigma^2}{\epsilon}$; on the other hand, as $r \to 2$ the second moments 
can be almost as small as $\sigma^2$. In general, $\sigmastar(\epsilon)$ is $\oo\p{\sigma \epsilon^{1/2-1/r}}$ if $r \in [1,2)$, is $\oo\big({\sigma \sqrt{\log(1/\epsilon)}}\big)$ if $r = 2$, and is $\oo\p{\sigma}$ if $r > 2$.

%% file: main-lp.tex
\section{Efficient Recovery Algorithms}
\label{sec:alg}

We now turn our attention to the question of efficient algorithms. 
The main point of this section is to prove Theorem~\ref{thm:alg-intro}, 
which yields efficient robust mean estimation for a general class of 
norms. 
%We will start by presenting our algorithm in the special case 
%of the $\ell_2$-norm as a warm-up.

\input recovery-l2

\subsection{General Case}
\label{sec:main-lp}

We are now ready to prove our general algorithmic result, 
Theorem~\ref{thm:alg-intro} from Section~\ref{sec:algo-intro}.
For convenience we recall Theorem~\ref{thm:alg-intro} here:
\begin{theorem*}
Suppose that $x_1, \ldots, x_n$ contains a subset $S$ of size $(1-\epsilon)n$ whose variance around 
its mean $\mu$ is bounded by $\sigma^2$ in the norm $\|\cdot\|$. 
Also suppose that Assumption~\ref{ass:opt} ($\kappa$-approximability)
holds for the dual norm $\|\cdot\|_*$. 
Then, if $\epsilon \leq \frac{1}{4}$, there is a polynomial-time algorithm whose output satisfies 
$\|\hat{\mu} - \mu\| = \oo\big(\sigma \sqrt{\kappa \epsilon}\big)$. 

If, in addition, $\|\cdot\|$ is $\gamma$-strongly convex, then even if $S$ 
only has size $\alpha n$ there is a polynomial-time algorithm such that 
$\|\hat{\mu} - \mu\| = \oo\big(\frac{\sqrt{\kappa} \sigma}{\sqrt{\gamma} \alpha}\big)$ 
with probability $\Omega(\alpha)$.
\end{theorem*}
Recall that bounded variance means that 
$\frac{1}{|S|} \sum_{i \in S} \langle x_i - \mu, v \rangle^2 \leq \sigma^2 \|v\|_*^2$ for 
all $v \in \bR^d$. There are two equivalent conditions to bounded variance that will be useful. 
The first 
is $\sup_{\|v\|_* \leq 1} v^{\top}\Sigma v \leq \sigma^2$ for all $v \in \bR^d$, 
where $\Sigma = \frac{1}{|S|} \sum_{i \in S} (x_i - \mu)(x_i - \mu)^{\top}$; this is 
useful because Assumption~\ref{ass:opt} allows us to $\kappa$-approximate this supremum 
for any given $\Sigma$. 

The second equivalent condition re-interprets $\sigma$ in terms of a matrix norm. 
Let $\|\cdot\|_{\psi}$ denote the norm $\|\cdot\|$ above, 
%(for which $\|\cdot\|_*$ is the dual norm), 
and for a matrix $M$ define the induced
$2\! \to\! \psi$-norm $\|M\|_{2 \to \psi}$ as $\sup_{\|u\|_2 \leq 1} \|Mu\|_{\psi}$. 
Then the set $S$ has variance at most $\sigma^2$ if and only if 
$\|[x_i - \mu]_{i \in S}\|_{2 \to \psi} \leq \sqrt{|S|}\sigma$.
This will be useful because induced norms satisfy helpful compositional properties such 
as $\|AB\|_{2 \to \psi} \leq \|A\|_{2 \to \psi}\|B\|_{2}$.

%main theorem on robust mean estimation 
%in $\ell_p$-norms. 
%The proof consists of combining the powering up results in Section~\ref{sec:powering-up} 
%(Propositions~\ref{prop:powering-up-lp}, \ref{prop:powering-up-lp-eps}) with a 
%generalization of Proposition~\ref{prop:recovery-l2} to $\ell_p$-norms. 
%We first give the generalization of Proposition~\ref{prop:recovery-l2}:
%\begin{proposition}
%\label{prop:recovery-lp}
%Let $x_1, \ldots, x_n \in \bR^d$, and let $S$ be a 
%set of size $\alpha n$ with mean $\mu$ and $2$nd moments bounded by $\sigma^2$ in $\ell_q$-norm: 
%$\|[x_i - \mu]_{i \in S}\|_{2 \to p} \leq \sigma\sqrt{|S|}$. 
%Then there is an efficient randomized algorithm %(Algorithm~\ref{alg:recover-lp}) 
%which with 
%probability $\Omega(\alpha)$ outputs a parameter $\hat{\mu}$ such that 
%$\|\mu - \hat{\mu}\|_p = \oo\Big({\frac{\sigma}{\alpha\sqrt{p-1}}}\Big)$. Moreover, if 
%$\alpha \geq \frac{3}{4}$ then %the algorithm outputs a parameter such that 
%$\|\mu - \hat{\mu}\|_p = \oo\p{\sigma\sqrt{1-\alpha}}$ with probability $1$.
%%\todo{check if $\mu$ must be the mean of $S_0$}
%\end{proposition}
The algorithm establishing Theorem~\ref{thm:alg-intro} 
is almost identical to Algorithm~\ref{alg:recover-l2}, with two changes. 
The first change is that on line \ref{line:if}, 
the quantity $4n\sigma^2$ is replaced with $4\kappa n \sigma^2$, where 
$\kappa$ is the approximation factor in Assumption~\ref{ass:opt}.
The second change is that in the optimization \eqref{eq:opt-2}, 
the constraint $Y \succeq 0, \tr(Y) \leq 1$ is replaced with 
$Y \in \sP$, where $\sP$ is the feasible set in Assumption~\ref{ass:opt}. 
In other words, the only difference is that rather than finding the 
maximum eigenvalue, we $\kappa$-approximate the 
$2\to \psi$ norm using Assumption~\ref{ass:opt}.
%\|\diag(Y)\|_{q/2} \leq 1$ (note these 
%are equal when $q = 2$). 
We therefore end up solving the saddle point problem
\begin{equation}
\label{eq:opt-lp}
%\max_{\substack{Y \succeq 0, \\ \|\diag(Y)\|_{q/2} \leq 1}} \min_{\substack{0 \leq W_{ji} \leq \frac{4-\alpha}{\alpha(2+\alpha) n}, \\ \sum_j W_{ji} = 1}} \sum_{i \in \sA} c_i (x_i - X_{\sA}w_i)^{\top}Y(x_i - X_{\sA}w_i).
\max_{Y \in \sP} \min_{\substack{0 \leq W_{ji} \leq \frac{4-\alpha}{\alpha(2+\alpha) n}, \\ \sum_j W_{ji} = 1}} \sum_{i \in \sA} c_i (x_i - X_{\sA}w_i)^{\top}Y(x_i - X_{\sA}w_i).
\end{equation}
Standard optimization algorithms such as Frank-Wolfe allow us to solve 
\eqref{eq:opt-lp} to any given precision with a polynomial number of calls to the linear 
optimization oracle guaranteed by Assumption~\ref{ass:opt}.

While Algorithm~\ref{alg:recover-l2} essentially minimizes the quantity $\|X-XW\|_{2}^2$, 
this new algorithm can be thought of as minimizing $\|X-XW\|_{2 \to \psi}^2$.
%, or equivalently $\|(X(I-W))^{\top}\|_{q \to 2}^2$.
%which is the squared operator norm when mapping from $\ell_2(\bR^n)$ to $\ell_p(\bR^d)$. 
However, for general norms computing the $2 \to \psi$ norm is NP-hard, and so 
we rely on a $\kappa$-approximate solution by optimizing over $\sP$.
%%%  to compute exactly 
%%% %(e.g. it is the dual of the 
%%% %cut-norm when $p = 1$). 
%%% for $p \in [1,2)$ \citep{steinberg2005computation}.
%%% The optimization \eqref{eq:opt-lp} therefore 
%%% employs a relaxation of the $2 \to p$ norm. 
%%% The result below, which follows from 
%%% Theorem 3 of \citet{nesterov1998semidefinite}, asserts this: %\todo{possibly save a line here}
%%% \begin{theorem}[\citeauthor{nesterov1998semidefinite}]
%%% \label{thm:grothendieck}
%%% For a matrix $A \in \bR^{d \times n}$, let $f(A) = \|A\|_{2 \to p}^2 = \max_{\|v\|_q \leq 1} \sum_{i=1}^n \langle a_i, v \rangle^2$, and let $g(A) = \max_{Y \succeq 0, \|\diag(Y)\|_{q/2} \leq 1} \sum_{i=1}^n a_i^{\top}Ya_i$. Then $f(A) \leq g(A) \leq \frac{\pi}{2} f(A)$.
%%% \end{theorem}
%%% Note that the first inequality is trivial since $vv^{\top}$ is a feasible value of $Y$; 
%%% the second inequality is established using a generalization of Grothendieck's inequality.
We are now ready to prove Theorem~\ref{thm:alg-intro}.

%\begin{algorithm}[b!]
%\caption{Algorithm for recovering the mean of a set with bounded $2$nd moments in $\ell_2$-norm.}
%\label{alg:recover-l2}
%\begin{algorithmic}[1]
%\STATE Initialize $c_i = 1$ for all $i=1,\ldots,n$ and $\sA = \{1,\ldots,n\}$.
%  \STATE Let $Y \in \bR^{d \times d}$ and $W \in \bR^{\sA \times \sA}$ 
%  be the maximizer/minimizer of the saddle point problem
%\label{line:begin-loop}
%  \begin{equation}
%  \label{eq:opt-lp-2}
%  \max_{\substack{Y \succeq 0, \\ \|\diag(Y)\|_{q/2} \leq 1}} \min_{\substack{0 \leq W_{ij} \leq \frac{4-\alpha}{\alpha(2+\alpha) n}, \\ \sum_i W_{ij} \leq 1}} \sum_{i \in \sA} c_i (x_i - X_{\sA}w_i)^{\top}Y(x_i - X_{\sA}w_i).
%  \end{equation}
%  \IF{$\sum_{i \in \sA} c_i(x_i - X_{\sA}w_i)^{\top}Y(x_i - X_{\sA}w_i) > 4n\sigma^2$} \label{line:if}
%    \STATE Let $\tau_i = (x_i - X_{\sA}w_i)^{\top}Y(x_i - X_{\sA}w_i)$ and $\tau_{\max} = \max_{i \in \sA} \tau_i$.
%    \STATE For $i \in \sA$, replace $c_i$ with $\p{1-\frac{\tau_i}{\tau_{\max}}}c_i$.
%    \STATE For all $i$ with $c_i < \frac{1}{2}$, remove $i$ from $\sA$.
%    \STATE Go back to line \ref{line:begin-loop}.
%  \ENDIF
%%\ENDWHILE
%\STATE Let $U\Lambda V^{\top}$ be the singular value decomposition of $W$.
%\STATE Let $\Lambda_0$ be the result of zeroing out all singular values less than $0.9$. %\frac{1}{2}$.
%\STATE Let $Z = X_{\sA}U\Lambda_0 V^{\top}(I - U(\Lambda - \Lambda_0)V^{\top})^{-1}$.
%\IF{$\rank(Z) = 1$}
%\STATE Output the average of the columns of $X_{\sA}$.
%\ELSE
%\STATE Output a column of $Z$ at random.
%\ENDIF
%\end{algorithmic}
%\end{algorithm}

\begin{prooff}{Theorem~\ref{thm:alg-intro}}
The proof is similar to Proposition~\ref{prop:recovery-l2}, so we only provide a 
sketch of the differences. 
First, the condition of Lemma~\ref{lem:invariant} still holds, now with 
$a$ equal to $\kappa n\sigma^2$ rather than $n\sigma^2$ 
due to the approximation ratio $\kappa$. %in Theorem~\ref{thm:grothendieck}.
(This is why we needed to change line~\ref{line:if}.) 

Next, we need to modify equations (\ref{eq:op-begin}-\ref{eq:xz-bound}) to hold for 
the $2 \to \psi$ norm rather than operator norm: %We have
\begin{align}
\|X_{\sA} - Z\|_{2 \to \psi} &= \|X_{\sA}(I-W)(I -W_1)^{-1}\|_{2 \to \psi} \\ % U(\Lambda - \Lambda_0)V^{\top})^{-1}\|_{q \to 2} \\
 &\stackrel{(i)}{\leq} \|X_{\sA}(I-W)\|_{2 \to \psi}\|(I - W_1)^{-1}\|_{2 \to 2} \\
 &\leq 10\|X_{\sA}(I-W)\|_{2 \to \psi} \\
 &\leq 10\sqrt{2}\|X_{\sA}(I-W)\diag(c_{\sA})^{1/2}\|_{2 \to \psi} \leq 20\sqrt{2\kappa n}\sigma.
\end{align}
Here (i) is from the general fact $\|AB\|_{2 \to \psi} \leq \|A\|_{2 \to \psi}\|B\|_{2 \to 2}$, 
and the rest of the inequalities follow for the same reasons as in 
(\ref{eq:op-begin}-\ref{eq:xz-bound}). %\todo{fix matrices and transposes for all of these norms}
%and (ii) is because all singular values of $U(\Lambda - \Lambda_0)V^{\top}$ are at most $0.9$.

We next need to modify equations (\ref{eq:fro-begin}-\ref{eq:fro-end}). This can be done 
with the following inequality:
\begin{lemma}
\label{lem:op-fro-p}
For any matrix $A = [a_1 \ \cdots \ a_n]$ of rank $r$ and any $\gamma$-strongly convex norm $\|\cdot\|_{\psi}$, 
%$p \in (1,2]$, 
we have 
$\sum_{i=1}^n \|a_i\|_{\psi}^2 \leq \frac{r}{\gamma} \|A\|_{2 \to \psi}^2$. 
%where $\frac{1}{p} + \frac{1}{q} = 1$ and $q \geq 2$.
%Moreover, if $r = 1$ then $\sum_{i=1}^n \|a_i\|_{\psi}^2 = \|A\|_{2 \to \psi}^2$.
\end{lemma}
This generalizes the inequality 
$\|A\|_F^2 \leq \rank(A)\cdot\|A\|_2^2$. Using Lemma~\ref{lem:op-fro-p} (proved below), we have
\begin{align}
\sum_{i \in S \cap \sA} \|z_i - \mu\|_{\psi}^2 
 &\leq \tfrac{\rank(Z)+1}{\gamma}\|[z_i - \mu]_{i \in S \cap \sA}\|_{2 \to \psi}^2 \\
 &\leq \tfrac{\rank(Z)+1}{\gamma}\p{\|[z_i - x_i]_{i \in S \cap \sA}\|_{2 \to \psi} + \|[x_i - \mu]_{i \in S \cap \sA}\|_{2 \to \psi}}^2 \\
 &= \oo\Big({\tfrac{\kappa\sigma^2n}{\alpha \gamma}}\Big).
\end{align}
The inequalities again follow for the same reasons as before.
If we choose $z_i$ at random, 
with probability $\Omega(\alpha)$ we will output a $z_i$
with $\|z_i - \mu\|_{\psi} = \oo\p{\frac{\sigma\sqrt{\kappa}}{\alpha\sqrt{\gamma}}}$. 
This completes the first part of the proposition.

For the second part, by the same reasoning as before we obtain $\tilde{\mu}$ 
with $\|X_{\sA} - \tilde{\mu}\bi^{\top}\|_{2 \to \psi} = \oo\p{\sqrt{n \kappa}\sigma}$, 
which implies that $\sA$ is resilient %in $\ell_p$-norm with 
with $\sigmabest(\epsilon) = \oo\p{\sigma\sqrt{\kappa\epsilon}}$ for $\epsilon \leq \frac{1}{2}$. 
The mean of $\sA$ will 
therefore be within distance $\oo\p{\sigma\sqrt{\kappa\epsilon}}$ of $\mu$ as before, 
which completes the proof.
\end{prooff}
We finish by proving Lemma~\ref{lem:op-fro-p}.
\begin{prooff}{Lemma~\ref{lem:op-fro-p}}
Let $s \in \{-1,+1\}^n$ be a uniformly random sign vector. We will compare 
$\bE_s[\|As\|_{\psi}^2]$ in two directions. Let $P$ be the projection onto the span of $A$. 
On the one hand, we have 
$\|As\|_{\psi}^2 = \|APs\|_{\psi}^2 \leq \|A\|_{2 \to \psi}^2\|Ps\|_2^2$, 
and hence $\bE_s[\|As\|_{\psi}^2] \leq \bE_s[\|Ps\|_2^2]\|A\|_{2 \to \psi}^2 = \rank(A)\|A\|_{2 \to \psi}^2$. 
On the other hand, similarly to \eqref{eq:cotype-induction} we have 
$\bE_s[\|As\|_{\psi}^2] \geq (1/\gamma)\sum_{i=1}^n \|a_i\|_{\psi}^2$ by the strong convexity of 
the norm $\|\cdot\|_{\psi}$. 
Combining these yields the desired result.
\end{prooff}
%By putting together the preceding results, we obtain Theorem~\ref{thm:main-lp}. 
%See Section~\ref{sec:main-lp-proof} for details.

%% file: recovery-l2.tex
%\vspace*{-2mm}

\subsection{Warm-Up: Recovery in $\ell_2$-norm}
\label{sec:main-l2}
We first prove a warm-up to Theorem~\ref{thm:alg-intro} which focuses on 
the $\ell_2$-norm.
% and assumes that the data has bounded second moments. 
%This assumption is stronger than resilience, e.g. second moments bounded by 
%$\sigma^2$ implies $\sigmabest(\epsilon) \leq \sqrtt{\frac{\epsilon}{1-\epsilon}}\sigma$; 
%see Section~\ref{sec:resilience-properties}. 
Our warm-up result is:
\begin{proposition}
\label{prop:recovery-l2}
Let $x_1, \ldots, x_n \in \bR^d$, and let $S$ be a 
subset of size $\alpha n$ with bounded variance in the $\ell_2$-norm:
$\lambda_{\max}(\frac{1}{|S|}\sum_{i \in S} (x_i - \mu)(x_i - \mu)^{\top}) \leq \sigma^2$, 
%\|[x_i - \mu]_{i \in S}\|_2 \leq \sigma \sqrt{|S|}$, 
where $\mu$ is the mean of $S$.
%$\frac{1}{|S|} \sum_{i \in S} (x_i - \mu)(x_i - \mu)^{\top} \preceq \sigma^2 I$, where 
%$\mu$ is the mean of $S$. 
Then there is an efficient randomized algorithm (Algorithm~\ref{alg:recover-l2}) which with 
probability $\Omega(\alpha)$ outputs a parameter $\hat{\mu}$ such that 
$\|\mu - \hat{\mu}\|_2 = \oo\p{\frac{\sigma}{\alpha}}$. Moreover, if 
$\alpha = 1-\epsilon \geq \frac{3}{4}$ then 
%the algorithm outputs a parameter such that 
$\|\mu - \hat{\mu}\|_2 = \oo\p{\sigma\sqrt{\epsilon}}$ with probability $1$.
%\todo{check if $\mu$ must be the mean of $S$}
\end{proposition}
At the heart of Algorithm~\ref{alg:recover-l2} is the following optimization problem:
\begin{align}
\notag \minim{W \in \bR^{n \times n}} & \opsep \|X-XW\|_{2}^2 \\ \label{eq:opt-1}
\subjto & \opsep 0 \leq W_{ji} \leq \frac{1}{\alpha n} \fasep \forall i,j, \quad \sum_j W_{ji} = 1 \fasep \forall i.
\end{align}
Here $X \in \bR^{d \times n}$ is the data matrix $[x_1 \ \cdots \ x_n]$ and 
$\|X-XW\|_2$ is the operator norm (maximum singular value) of $X-XW$. 
Note that \eqref{eq:opt-1} can be expressed as a semidefinite program; however, it can 
actually be solved more efficiently than this, via a singular value decomposition 
(see Section~\ref{sec:unitary-proof} for details).
%and $\|\cdot\|_{2}$ denotes operator norm. 

The idea behind \eqref{eq:opt-1} is to re-construct each $x_i$ 
as an average of $\alpha n$ other $x_j$. Note that by assumption we can 
always re-construct each element of $S$ using the mean of $S$, and have small error.
Intuitively, any element that cannot be re-constructed well must not lie in 
$S$, and can be safely removed.
% conversely, even if a point lies outside $S$, if it 
%can be re-constructed well then it will not end up perturbing the optimal $XW$ by too much.
%\todo{how understandable is this intuition?}
We do a soft form of removal by maintaining weights $c_i$ on the points $x_i$ 
(initially all $1$), and downweighting points with high reconstruction error.
We also maintain an active set $\sA$ of points with $c_i \geq \frac{1}{2}$.

Informally, Algorithm~\ref{alg:recover-l2} for estimating $\mu$ takes the following form:
\begin{enumerate}
\setlength\itemsep{0em}
\item Solve the optimization problem \eqref{eq:opt-1}.
\item If the optimum is $\gg \sigma^2 n$, then find the 
      columns of $X$ that are responsible for the optimum being 
      large, and downweight them. 
\item Otherwise, if the optimum is $\oo(\sigma^2n)$, then take a 
      low rank approximation $W_0$ to $W$, and 
      return a randomly chosen column of $XW_0$.
\end{enumerate}
The hope in step $3$ is that the low rank projection $XW_0$ will be close 
to $\mu$ for the columns belonging to $S$.
The choice of operator norm is crucial: it means we can actually expect $XW$ to be 
close to $X$ (on the order of $\sigma\sqrt{n}$). In contrast, the Frobenius 
norm would scale as $\sigma \sqrt{nd}$.

\input alg-recover-l2

Finally, we note that $\|X-XW\|_{2}^2$ is equal to
\begin{equation}
\|X-XW\|_{2}^2 = 
\max_{Y \succeq 0, \tr(Y) \leq 1} \sum_{i=1}^n (x_i - Xw_i)^{\top}Y(x_i - Xw_i), 
\end{equation}
which is the form we use in Algorithm~\ref{alg:recover-l2}. 

% In step 2 above, we will decrease the weights $c_i$ of points that 
%cause the optimum to be too large; we can think of this as a soft outlier removal step.
%\todo{save space here?}

\begin{prooff}{Proposition~\ref{prop:recovery-l2}}
We need to show two things: (1) that the outlier removal step removes 
many more outliers than good points, and (2) that many columns of $XW_0$
are close to $\mu$.

\paragraph{Outlier removal.}
To analyze the outlier removal step 
%Before proving Proposition~\ref{prop:recovery-l2}, we provide 
%a rough outline. First, we need to show that when we downweight outliers 
(step 2 above, or lines \ref{line:outlier-begin}-\ref{line:outlier-end} of 
Algorithm~\ref{alg:recover-l2}), 
we make use of the following general lemma:
%most of the points removed are not in $S$. This is formalized as follows:
\begin{lemma}
\label{lem:invariant}
%The following invariants hold throughout the algorithm:
%(1) $|S \cap \sA| \geq \frac{\alpha(2+\alpha)}{4-\alpha} n$, and (2)
%$\sum_{i \in S} (1-c_i) \leq \frac{\alpha}{4}\sum_{i=1}^n (1-c_i)$.
For any scalars $\tau_i$ and $a$, suppose that $\sum_{i \in \sA} c_i\tau_i \geq 4a$ 
while $\sum_{i \in S \cap \sA} c_i\tau_i \leq \alpha a$.
Then the following invariants are preserved by lines 
\ref{line:outlier-begin}-\ref{line:outlier-end} of Algorithm~\ref{alg:recover-l2}:
(i) $\sum_{i \in S} (1-c_i) \leq \frac{\alpha}{4}\sum_{i=1}^n (1-c_i)$,
and (ii) $|S \cap \sA| \geq \frac{\alpha(2+\alpha)}{4-\alpha} n$.
\end{lemma}
Lemma~\ref{lem:invariant} says that we downweight points within $S$ at least $4$ times 
slower than we do
overall (property i), and in particular we never remove too many points from $S$ 
(property ii). This type of lemma is not new (cf.~Lemma 4.5 of \citet{charikar2017learning}) 
but for completeness we prove it in Section~\ref{sec:invariant-proof}.

We will show that we can take $a = n\sigma^2$ in Lemma~\ref{lem:invariant}, or in 
other words that 
$\sum_{i \in S \cap \sA} c_i\tau_i^{\star} \leq \alpha n\sigma^2$.
Let $\tau_i(w) = (x_i - X_{\sA}w)^{\top}(x_i - X_{\sA}w)$, and note that 
$\tau_i^{\star} = \tau_i(w_i) = \min\{\tau_i(w) \mid 0 \leq w_j \leq \frac{1}{\alpha n}, \sum_j w_j = 1\}$. This is because for a fixed $Y$, each of the $w_i$ are optimized independently. 

We can therefore bound $\tau_i^{\star}$ by substituting any feasible $\hat{w}_i$. We will choose 
$\hat{W}_{ji} = \frac{\bI[j \in S \cap \sA]}{|S \cap \sA|}$, in which case 
$X_{\sA}\hat{w}_i = \hat{\mu}$, where $\hat{\mu}$ is the average of $x_j$ over $S \cap \sA$. 
Then we have
\begin{align}
%\sum_{i \in S \cap \sA} c_i(x_i - X_{\sA}w_i)^{\top}Y(x_i - X_{\sA}w_i)
\sum_{i \in S \cap \sA} c_i\tau_i^{\star}
\label{eq:resilient-sum}
 &\leq \sum_{i \in S \cap \sA} c_i\tau_i(\hat{w}_i) \\
 &\leq \sum_{i \in S \cap \sA} c_i(x_i - \hat{\mu})^{\top}Y(x_i - \hat{\mu}) \\
 &\stackrel{(i)}{\leq} \sum_{i \in S \cap \sA} c_i(x_i - \mu)^{\top}Y(x_i - \mu) \\
 &\leq \sum_{i \in S} (x_i - \mu)^{\top}Y(x_i - \mu) 
 \leq \alpha n \sigma^2 \tr(Y) \leq \alpha n \sigma^2
\end{align}
as desired;
(i) is because the covariance around the mean ($\hat{\mu}$) is smaller than 
around any other point ($\mu$). %\todo{save space here?}

\paragraph{Analyzing $XW_0$.}
By Lemma~\ref{lem:invariant}, we will eventually exit the if statement and obtain 
$Z = X_{\sA}W_0$. It therefore remains to analyze $Z$; we will show in particular 
that $\|Z_{\sA \cap S} - \mu \bi^{\top}\|_F$ is small, where the subscript indicates 
restricting to the columns in $S \cap \sA$.
%$XW_0$. We will show that on the columns in $S \cap \sA$, 
%the Frobenius norm $\|XW_0 - \mu \bi^{\top}\|_F$ is small. 
%Note that 
%$\|XW_0 - \mu \bi^{\top}\|_F \leq \sqrt{\rank(W_0)}\|XW_0 - \mu \bi^{\top}\|_2$ 
%and $\|XW_0 - \mu \bi^{\top}\|_2 \leq \|XW_0 - X\|_2 + \|X - \mu \bi^{\top}\|_2$. 
%It thus suffices to show that $W_0$ has small rank, and that 
%$XW_0$ is close to $X$ in spectral norm 
%(we already know that $X$ and $\mu\bi^{\top}$ are close by assumption).
At a high level, it suffices to show that $W_0$ has low rank (so that Frobenius 
norm is close to spectral norm) and that $XW_0$ and $X$ are close in spectral 
norm (note that $X$ and $\mu\bi^{\top}$ are close by assumption). %, at least within $S$).

%This ensures that we eventually find a $W$ such that $\|X - XW\|_2$ is small. 
%Then, we show that $W$ can be approximated by a low rank matrix $W_0$, such that 
%$\|X - XW_0\|_2$ is also small. Since $XW_0$ is close to $X$ and $X$ is close to $\mu \bi^{\top}$ 
%by assumption, $XW_0$ is close to $\mu \bi^{\top}$ 
%(in operator norm, and hence also in Frobenius norm since $XW_0$ is low rank). 
%This allows us to conclude that a randomly chosen column of $XW_0$ is close to $\mu$, as 
%desired. We proceed to show this formally.

%First, the following invariants are satisfied by 
%Algorithm~\ref{alg:recover-l2}:
%We establish these later in this section. Taking 
%Lemma~\ref{lem:invariant} as given for now, we proceed 
%to analyze the matrix $Z$. 
To bound $\rank(W_0)$, note that the constraints in \eqref{eq:opt-2} 
imply that $\|W\|_F^2 \leq \frac{4-\alpha}{\alpha(2+\alpha)}$, and so at most 
$\frac{4-\alpha}{0.81\alpha(2+\alpha)}$ singular values of $W$ can be greater 
than $0.9$. Importantly, at most $1$ singular value can be greater than $0.9$ 
if $\alpha \geq \frac{3}{4}$, and at most $\oo(\frac{1}{\alpha})$ can be in general. 
Therefore, $\rank(W_0) \leq \oo(\frac{1}{\alpha})$.

Next, we show that $X_{\sA}$ and $Z$ are close in operator norm. 
Indeed, $X_{\sA} - Z = X_{\sA}(I-W_0) = X_{\sA}(I-W)(I-W_1)^{-1}$, hence:
\begin{align}
\label{eq:op-begin}
\|X_{\sA}-Z\|_{2}
 &= \|X_{\sA}(I-W)(I-W_1)^{-1}\|_2 \\
 &\leq \|X_{\sA}(I-W)\|_2 \|(I-W_1)^{-1}\|_2 \\
% - X_{\sA}U\Lambda_0V^{\top}(I - U(\Lambda - \Lambda_0) V^{\top})^{-1}\|_{2} \\
% &\leq \|X_{\sA}(I - U(\Lambda - \Lambda_0)V^{\top}) - X_{\sA}U\Lambda_0V^{\top}\|_{2}\|(I - U(\Lambda - \Lambda_0)V^{\top})^{-1}\|_{2} \\
 &\stackrel{(i)}{\leq} 10\|X_{\sA}(I - W)\|_2 \\
%U(\Lambda - \Lambda_0)V^{\top}) - X_{\sA}U\Lambda_0V^{\top}\|_{2} \\
% &= 10\|X_{\sA}(I - U\Lambda V^{\top})\|_{2} \\
% &= 10\|X_{\sA}(I - W)\|_{2} \\
 &\stackrel{(ii)}{\leq} 10\sqrt{2}\|X_{\sA}(I-W)\diag(c_{\sA})^{1/2}\|_{2} 
 \stackrel{(iii)}{\leq} 20\sqrt{2n}\sigma.
\label{eq:xz-bound}
\end{align}
Here (i) is because all singular values of $W_1$ are less than 
$0.9$, (ii) is because $\diag(c_{\sA})^{1/2} \succeq \frac{1}{\sqrt{2}}I$, 
and (iii) is by the condition in the if statement (line \ref{line:if} of Algorithm~\ref{alg:recover-l2}), since the sum on line \ref{line:if} 
is equal to $\|X_{\sA}(I-W)\diag(c_{\sA})^{1/2}\|_2^2$.

Combining the previous two observations, we have 
\begin{align}
\label{eq:fro-begin}
\sum_{i \in S \cap \sA} \|z_i - \mu\|_2^2 
% &= \|[z_i - \mu]_{i \in S \cap \sA}\|_F^2 \\
 &\leq (\rank(Z)+1)\|[z_i - \mu]_{i \in S \cap \sA}\|_{2}^2 \\
 &\leq (\rank(Z)+1)\p{\|[z_i - x_i]_{i \in S \cap \sA}\|_{2} + \|[x_i - \mu]_{i \in S \cap \sA}\|_{2}}^2 \\
 &\stackrel{(i)}{\leq} (\rank(Z)+1)\p{20\sqrt{2n}\sigma + \sqrt{\alpha n}\sigma}^2 
 = \oo\p{\tfrac{\sigma^2}{\alpha}n}.
\label{eq:fro-end}
\end{align}
Here (i) uses the preceding bound on $\|X_{\sA} - Z\|_{2}$, together 
with the $2$nd moment bound %, which is equivalent to
$\|[x_i - \mu]_{i \in S}\|_{2} \leq \sqrt{\alpha n}\sigma$.
Note that $\rank(Z) \leq \rank(W_0) = \oo(\frac{1}{\alpha})$.

Since $|S \cap \sA| = \Omega(\alpha n)$ by Lemma~\ref{lem:invariant}, 
the average value of $\|z_i - \mu\|_2^2$ over $S \cap \sA$ is $\oo\big({\frac{\sigma^2}{\alpha^2}}\big)$, 
and hence with probability at least $\frac{|S \cap \sA|}{2|\sA|} = \Omega(\alpha)$, a randomly 
chosen $z_i$ will be within distance $\oo\p{\frac{\sigma}{\alpha}}$ of $\mu$, which completes
the first part of Proposition~\ref{prop:recovery-l2}.

For the second part, when $\alpha = 1-\epsilon \geq \frac{3}{4}$, recall that we have 
$\rank(W_0) = 1$, and that $W_0 = (W-W_1)(I-W_1)^{-1}$. 
%Now let us consider the matrix 
%$W_0 = U\Lambda_0V^{\top}(I - U(\Lambda - \Lambda_0)V^{\top})^{-1}$, which also has rank $1$. 
%Note that $\bi^{\top}W = \bi^{\top}$ by the constraints in \eqref{eq:opt-2}; we claim that 
%$\bi^{\top}W_0 = \bi^{\top}$ as well.
%If we let $W_1 = U(\Lambda - \Lambda_0)V^{\top}$, then we can write 
%$W_0 = (W - W_1)(I-W_1)^{-1}$. In particular, 
One can then verify that $\bi^{\top}W_0 = \bi^{\top}$.
%%% Also, %$W_0$ is stochastic, since
%%% $\bi^{\top}W_0 = (\bi^{\top}W - \bi^{\top}W_1)(I - W_1)^{-1} = \bi^{\top}(I-W_1)(I-W_1)^{-1} = \bi^{\top}$.
%
%So, $W_0$ is a rank-$1$ matrix satisfying $\bi^{\top}W_0 = \bi^{\top}$. This 
Therefore, $W_0 = u\bi^{\top}$ for some $u$. % satisfying $\bi^{\top}u = 1$. 
Letting $\tilde{\mu} = X_{\sA}u$,
we have $\|X_{\sA} - \tilde{\mu}\bi^{\top}\|_{2} \leq 20\sqrt{2n}\sigma$ 
by \eqref{eq:xz-bound}. In particular, 
$\sA$ is resilient (around its mean) %; see Section~\ref{sec:resilience-properties}) 
with $\sigma(\epsilon) \leq 20\sigma\sqrt{\frac{2\epsilon}{1-\epsilon}} \leq 40\sigma\sqrt{\epsilon}$ for $\epsilon \leq \frac{1}{2}$.
Thus by the proof of Proposition~\ref{prop:resilience} and the fact 
that $|\sA| \geq |S \cap \sA| \geq \frac{\alpha (2+\alpha)}{4-\alpha} n \geq (1-\frac{5}{3}(1-\alpha))n$, 
the mean of $\sA$ is within $\oo(\sigma\sqrt{1-\alpha})$ of $\mu$, as desired.
\end{prooff}

%% file: alg-recover-l2.tex
\begin{algorithm}[t!]
\caption{Algorithm for recovering the mean of a set with bounded variance in $\ell_2$-norm.}
\label{alg:recover-l2}
\begin{algorithmic}[1]
\STATE Initialize $c_i = 1$ for all $i=1,\ldots,n$ and $\sA = \{1,\ldots,n\}$.
  \STATE Let $Y \in \bR^{d \times d}$ and $W \in \bR^{\sA \times \sA}$ 
  be the maximizer/minimizer of the saddle point problem
\label{line:begin-loop}
%  \vskip -0.04in
  \begin{equation}
  \label{eq:opt-2}
  \max_{\substack{Y \succeq 0, \\ \tr(Y) \leq 1}} \min_{\substack{0 \leq W_{ji} \leq \frac{4-\alpha}{\alpha(2+\alpha) n}, \\ \sum_j W_{ji} = 1}} \sum_{i \in \sA} c_i (x_i - X_{\sA}w_i)^{\top}Y(x_i - X_{\sA}w_i).
  \end{equation}
%  \vskip -0.10in \phantom{x}
  \STATE Let $\tau_i^{\star} = (x_i - X_{\sA}w_i)^{\top}Y(x_i - X_{\sA}w_i)$.
  %\IF{$\sum_{i \in \sA} c_i(x_i - X_{\sA}w_i)^{\top}Y(x_i - X_{\sA}w_i) > 4n\sigma^2$} 
  \IF{$\sum_{i \in \sA} c_i \tau_i^{\star} > 4n\sigma^2$} 
\protect\label{line:if}
%    \STATE Let $\tau_i = (x_i - X_{\sA}w_i)^{\top}Y(x_i - X_{\sA}w_i)$.% and $\tau_{\max} = \max_{i \in \sA} \tau_i$.
    \STATE For $i \in \sA$, replace $c_i$ with $\p{1-\frac{\tau_i^{\star}}{\tau_{\max}}}c_i$, where $\tau_{\max} = \max_{i \in \sA} \tau_i^{\star}$.
\label{line:outlier-begin}
    \STATE For all $i$ with $c_i < \frac{1}{2}$, remove $i$ from $\sA$.
\label{line:outlier-end}
    \STATE Go back to line \ref{line:begin-loop}.
  \ENDIF 
\protect\label{line:endif}
%\ENDWHILE
%\STATE Let $U\Lambda V^{\top}$ be the singular value decomposition of $W$.
%\STATE Let $\Lambda_0$ be the result of zeroing out all singular values less than $0.9$. %\frac{1}{2}$.
\STATE Let $W_1$ be the result of zeroing out all singular values of $W$ that are greater than $0.9$.
\STATE Let $Z = X_{\sA}W_0$, where $W_0 = (W-W_1)(I-W_1)^{-1}$. %U\Lambda_0 V^{\top}(I - U(\Lambda - \Lambda_0)V^{\top})^{-1}$.
\IF{$\rank(Z) = 1$}
\STATE Output the average of the columns of $X_{\sA}$.
\ELSE
\STATE Output a column of $Z$ at random.
\ENDIF
\end{algorithmic}
\end{algorithm}

%% file: rank-k.tex
\section{Robust Low-Rank Recovery}
\label{sec:rank-k}

In this section we present results on rank-$k$ recovery. We first 
justify the definition of rank-resilience (Definition~\ref{def:resilience-rank-k}) 
by showing that it is information-theoretically sufficient for (approximately) recovering
the best rank-$k$ subspace. Then, we provide an algorithm showing that this subspace can 
be recovered efficiently.
%\todo{save line here?}

\subsection{Information-Theoretic Sufficiency}

%We start with the following 
%lemma which is core to information-theoretic recoverability:
%\begin{lemma}
%\label{lem:rank-k-inv}
%Suppose that $S$ is rank $k$ $(\sigma,\delta)$-resilient and let $T \subseteq S$ 
%with $|T| \geq (1-\delta)|S|$. Furthermore let $P$ be any rank $k$ projection 
%matrix such that $\|(I-P)X_T\|_{2} \leq \sigma_0\sqrt{|S|}$. Then 
%$\|(I-P)X\|_{2} \leq 2\sigma_0\sqrt{|S|}$.
%\end{lemma}
%\begin{proof}[Proof of Lemma~\ref{lem:rank-k-inv}]
%\end{proof}
%Lemma~\ref{lem:rank-k-inv} says that for any large subset $T$ of $S$, if we can 
%find a projection $P$ that represents most of the variation on $T$, 
%then it also represents most of the variation on all of $S$. (Note that the 
%converse, moving from $S$ down to $T$, is trivially true.) In general, such equivalences 
%between a property on a set and on all of its large subsets will be sufficient for 
%information-theoretic recoverability.

Let $X_S = [x_i]_{i \in S}$. 
Recall that $\delta$-rank-resilience 
%Definition~\ref{def:resilience-rank-k} 
asks that $\col(X_T) = \col(X_S)$ and 
$\|X_T^{\dagger}X_S\|_2 \leq 2$ 
for $|T| \geq (1-\delta)|S|$. This is justified by the following: % result:
\begin{proposition}
\label{prop:info-theory-rank-k}
Let $S \subseteq [n]$ be a set of points of size $(1-\delta)n$ 
that is $\frac{\delta}{1-\delta}$-rank-resilient. Then it is possible to 
output a rank-$k$ projection matrix $P$ such that 
$\|(I-P)X_S\|_{2} \leq 2\sigma_{k+1}(X_S)$.
\end{proposition}
\begin{proof}
Find the $\frac{\delta}{1-\delta}$-rank-resilient set $S'$ of size $(1-\delta)n$ 
such that $\sigma_{k+1}(X_S')$ is smallest, and let $P$ be the projection 
onto the top $k$ singular vectors of $X_{S'}$. Then we have 
$\|(I-P)X_{S'}\|_{2} = \sigma_{k+1}(X_{S'}) \leq \sigma_{k+1}(X_S)$.
Moreover, if we let $T = S \cap S'$, we have 
$\|(I-P)X_T\|_{2} \leq \|(I-P)X_{S'}\|_{2} \leq \sigma_{k+1}(X_S)$ as well.
By pigeonhole, $|T| \geq (1-2\delta)n = ({1 - \frac{\delta}{1-\delta}})|S|$. 
Therefore $\col(X_T) = \col(X_S)$, and
$\|(I-P)X_S\|_{2} = \|(I-P)X_TX_T^{\dagger}X_S\|_{2} \leq \|(I-P)X_T\|_{2}\|X_T^{\dagger}X_S\|_{2} \leq 2\sigma_{k+1}(X_S)$ as claimed.
\end{proof}
%\todo{should we put forth alternative definitions and provide counterexamples? or just stick with this one?}

\vskip -0.45in
\phantom{x}

\subsection{Efficient Recovery}
%Having established the possibility of recovering $P$ 
%information-theoretically, we next provide an efficient 
%algorithm for doing so. 
We now address the question of efficient recovery, by proving 
Theorem~\ref{thm:main-rank-k}, whose statement we recall here (with a few details added). %recall here. 
For convenience let $\sigma \eqdef \sigma_{k+1}(X_S)/\sqrt{|S|}$.
%We recall the statement of Theorem~\ref{thm:main-rank-k}:
\begin{theorem*}
%\label{thm:recovery-rank-k}
Let $\delta \leq \frac{1}{3}$. 
If a set of $n$ points contains a subset $S$ of size 
$(1-\delta)n$ that is $\delta$-rank-resilient, 
then there is an efficient algorithm (Algorithm~\ref{alg:recover-rank-k}) 
for recovering a matrix $P$ of rank at most $15k$
such that $\|(I-P)X_S\|_{2} = \oo(\sigma\sqrt{|S|}) = \oo(\sigma_{k+1}(X_S))$.
\end{theorem*}
Algorithm~\ref{alg:recover-rank-k} is quite similar to 
Algorithm~\ref{alg:recover-l2}, but we include it for completeness.
The proof is also similar to the proof of Proposition~\ref{prop:recovery-l2},
but there are enough differences that we provide details below.

\input alg-rank-k

\begin{prooff}{Theorem~\ref{thm:main-rank-k}} %recovery-rank-k}}
First, we show that Lemma~\ref{lem:invariant} holds with $a = 2n\sigma^2$. 
As before, the $q_i$ can be optimized independently for each $i$, so 
$\sum_{i \in S \cap \sA} c_i\tau_i^{\star} \leq \sum_{i \in S \cap \sA} c_i\tau_i(\hat{q}_i)$ for 
any matrix $\hat{Q}$. %is upper-bounded by its value at any $Q$. 
Take $\hat{Q}$ to be the projection onto the top $k$ (right) singular vectors of $X_{S \cap \sA}$. 
Then we have $\sum_{i \in S \cap \sA} \lambda \|\hat{q}_i\|_2^2 \leq \lambda \|\hat{Q}\|_F^2 = \lambda k$. 
We also have $\sum_{i \in S \cap \sA} c_i(x_i - X\hat{q}_i)^{\top}Y(x_i - X\hat{q}_i) \leq \sigma_{k+1}(X_{S \cap \sA})^2 \leq \sigma^2|S| \leq (1-\delta)n\sigma^2$. 
Together these imply that 
$\sum_{i \in S \cap \sA} c_i\tau_i(\hat{q}_i) \leq (1-\delta)n\sigma^2 + \lambda k = 2(1-\delta)n\sigma^2$, so we can indeed take $a = 2n\sigma^2$.
%
%In particular, this holds for the $p_i$ obtained from \eqref{eq:opt-rank-k}, since each of 
%the $p_i$ can be optimized independently for fixed $Y$.
%Consequently, the condition of Lemma~\ref{lem:invariant} holds with $a = 2n\sigma^2$, 
%and we thus have 
Consequently, 
$|S \cap \sA| \geq \frac{(1-\delta)(3-\delta)}{3+\delta}n \geq (1-\frac{5\delta}{3})n$.
%We claim that the following invariants hold, analogously to 
%Lemma~\ref{lem:invariant}: (1) $|S \cap \sA| \geq (1-\frac{5\delta}{3})n$; 
%(2) $\sum_{i \in S} (1-c_i) \leq \frac{1-\delta}{4}\sum_{i=1}^n (1-c_i)$.
%Indeed, suppose that the condition in the if statement (line \ref{line:if-rank-k}) holds. 
%Then $\sum_{i \in \sA} c_i \tau_i \geq 8n \sigma^2$ while 
%$\sum_{i \in S \cap \sA} \leq 2(1-\delta)n\sigma^2$. 
%By the same logic as in Lemma~\ref{lem:invariant}, $\sum_{i=1}^n (1-c_i)$ must increase 
%at least $\frac{1-\delta}{4}$ times as fast as $\sum_{i \in S} (1-c_i)$, which 
%establishes the invariant (1). The invariant (2) holds because 
%$\sum_{i \in S} (1-c_i) \leq \frac{1-\delta}{3+\delta}\sum_{i \not\in S} (1-c_i) \leq \frac{\delta}{3} n$, and hence $\#\{c_i \in S \mid c_i < \frac{1}{2}\} \leq \frac{2\delta}{3}n$.

Now, when we get to line \ref{line:post-if-rank-k}, we 
%have $|S \cap \sA| \geq (1-\frac{5\delta}{3})n$. Moreover, 
have $\|X_{\sA}(I-Q)\diag(c_{\sA})^{1/2}\|_2^2 + \lambda \|Q\|_F^2 \leq 8n\sigma^2$, 
and in particular each of the two terms individually is bounded by $8n\sigma^2$.
Therefore, $\|Q\|_F^2 \leq \frac{8n\sigma^2}{\lambda} = \frac{8k}{1-\delta}$,
and so $Q_0$ will have rank at most $\frac{8k}{0.81(1-\delta)} \leq 15k$. Moreover, 
$\|X_{\sA}(I-Q)\|_{2} \leq \sqrt{2}\|X_{\sA}(I-Q)\diag(c_{\sA})^{1/2}\|_{2} \leq 4\sqrt{n}\sigma$. 
We then have
\begin{align}
\|(I-P)X_{\sA}\|_2 
 &= \|(I-X_{\sA}Q_0X_{\sA}^{\dagger})X_{\sA}\|_{2} \\
 &= \|X_{\sA}(I - Q_0)X_{\sA}^{\dagger}X_{\sA}\|_2 \\
 &\leq \|X_{\sA}(I-Q_0)\|_2 \\
 &= \|X_{\sA}(I-Q)(I-Q_1)^{-1}\|_{2} \\
 &\leq 10\|X_{\sA}(I-Q)\|_{2} \leq 40\sqrt{n}\sigma = \oo\p{\sigma\sqrt{|S|}}.
\end{align}
By the same argument as Proposition~\ref{prop:info-theory-rank-k}, 
since $|S \cap \sA| \geq \frac{1-5\delta/3}{1-\delta}|S| \geq (1-\delta)|S|$, it 
follows that $\|(I-P)X_S\|_{2} \leq 2\|(I-P)X_{S \cap \sA}\|_{2} = \oo\p{\sigma\sqrt{|S|}}$ 
as well, which completes the proof.
\end{prooff}

%% file: alg-rank-k.tex
\begin{algorithm}
\caption{Algorithm for recovering a rank-$k$ subspace.}
\label{alg:recover-rank-k}
\begin{algorithmic}[1]
\STATE Initialize $c_i = 1$ for all $i=1,\ldots,n$ and $\sA = \{1,\ldots,n\}$. 
        Set $\lambda = \frac{(1-\delta)n\sigma^2}{k}$.
  \STATE Let $Y \in \bR^{d \times d}$ and $Q \in \bR^{\sA \times \sA}$ 
  be the maximizer/minimizer of the saddle point problem
  \begin{equation}
  \label{eq:opt-rank-k}
  \max_{\substack{Y \succeq 0, \\ \tr(Y) \leq 1}} \min_{Q \in \bR^{n \times n}} \sum_{i \in \sA} c_i [(x_i - X_{\sA}q_i)^{\top}Y(x_i - X_{\sA}q_i) + \lambda \|q_i\|_2^2].
  \end{equation}
  \STATE Let $\tau_i^{\star} = (x_i - X_{\sA}q_i)^{\top}Y(x_i - X_{\sA}q_i) + \lambda \|q_i\|_2^2$.
\label{line:begin-loop-rank-k}
  \IF{$\sum_{i \in \sA} c_i\tau_i^{\star} > 8n\sigma^2$} 
  %\IF{$\sum_{i \in \sA} c_i[(x_i - X_{\sA}q_i)^{\top}Y(x_i - X_{\sA}q_i) + \lambda \|q_i\|_2^2] > 8n\sigma^2$} 
\label{line:if-rank-k}
    %\STATE Let $\tau_i = (x_i - X_{\sA}q_i)^{\top}Y(x_i - X_{\sA}q_i) + \lambda \|q_i\|_2^2$.
    \STATE For $i \in \sA$, replace $c_i$ with $({1-\frac{\tau_i^{\star}}{\tau_{\max}}})c_i$, where $\tau_{\max} = \max_{i \in \sA} \tau_i^{\star}$.
    \STATE For all $i$ with $c_i < \frac{1}{2}$, remove $i$ from $\sA$.
    \STATE Go back to line \ref{line:begin-loop-rank-k}.
  \ENDIF
\STATE Let $Q_1$ be the result of zeroing out all singular values of $Q$ greater than $0.9$.
\label{line:post-if-rank-k}
\STATE Output $P = X_{\sA}Q_0X_{\sA}^{\dagger}$, where $Q_0 = (Q - Q_1)(I - Q_1)^{-1}$.
\end{algorithmic}
\end{algorithm}

%% file: concentration.tex
\section{Finite-Sample Concentration}
\label{sec:concentration}

In this section we provide two general finite-sample concentration 
results that establish resilience with high probability.
The first holds for arbitrary resilient distributions 
but has suboptimal sample complexity, while the latter is specialized to 
distributions with bounded variance and has near-optimal sample complexity.

\subsection{Concentration for Resilient Distributions}

Our first result, stated as Proposition~\ref{prop:concentration-resilient} in 
Section~\ref{sec:intro-concentration}, applies to any $(\sigma,\epsilon)$-resilient 
distribution $p$; recall that $p$ is $(\sigma,\epsilon)$-resilient iff 
$\|\bE[x \mid E] - \mu\| \leq \sigma$ for any event $E$ of probability $1-\epsilon$. 
%\todo{no longer true, fix this} equivalently (by Lemma~\ref{lem:reverse}), it is $(\sigma,\epsilon)$-resilient 
%iff $\|\bE[x \mid E'] - \mu\| \leq \frac{1-\epsilon}{\epsilon}\sigma$ for any event 
%$E'$ of probability $\epsilon$.

We define the \emph{covering number} of the unit ball in 
a norm $\|\cdot\|_*$ to be the minimum $M$ for which there 
are vectors $v_1, \ldots, v_M$, each with $\|v_j\|_* \leq 1$, such that 
$\max_{j=1}^M \langle x, v_j \rangle \geq \frac{1}{2}\sup_{\|v\|_* \leq 1} \langle x, v \rangle$ 
for all vectors $v \in \bR^d$. Note that $\log M$ is a measure 
of the effective dimension of the unit ball, i.e. 
$\log M = \Theta(d)$ if $\|\cdot\|_*$ is the $\ell_{\infty}$ or $\ell_2$ norm, 
while $\log M = \Theta(\log d)$ for the $\ell_1$ norm.

We recall Proposition~\ref{prop:concentration-resilient} for convenience:
\begin{proposition*}
%\label{prop:concentration-resilient}
Suppose that a distribution $p$ is $(\sigma,\epsilon)$-resilient around its mean 
$\mu$ with $\epsilon < \frac{1}{2}$.
%$\|\bE[x \mid E] - \mu\| \leq \sigma$ for any event 
%$E$ of probability at least $1-\epsilon$. 
Let $B$ be such that 
$\bP[\|x - \mu\| \geq B] \leq \epsilon/2$. Also let 
$M$ be the covering number of the unit ball in the dual norm 
$\|\cdot\|_*$. % to $\|\cdot\|$.

Then, given 
$n$ samples $x_1, \ldots, x_n \sim p$, with probability 
$1 - \delta - \exp(-\epsilon n / 6)$ there is a subset $T$ of 
$(1-\epsilon)n$ of the $x_i$ such that $T$ is 
$(\sigma', \epsilon)$-resilient with 
$\sigma' = \oo\bigg(\sigma \Big(1 + \sqrt{\frac{\log(M/\delta)}{\epsilon^2 n}} + \frac{(B/\sigma) \log(M/\delta)}{n}\Big)\bigg)$.
\end{proposition*}

\begin{proof}
Let $p'$ be the distribution of samples from $p$ conditioned on 
$\|x - \mu\| \leq B$. Note that $p'$ is $(\sigma, \frac{\epsilon}{2})$-resilient 
since every event with probability $1-\epsilon/2$ in $p'$ is an event of 
probability $(1-\epsilon/2)^2 \geq 1-\epsilon$ in $p$. Moreover, with probability 
$1 - \exp(-\epsilon n/6)$, at least $(1-\epsilon)n$ of the samples from 
$p$ will come from $p'$. Therefore, we can focus on establishing resilience 
of the $n' = (1-\epsilon)n$ samples from $p'$.
%\todo{indexing shifts here}

With a slight abuse of notation, let $x_1, \ldots, x_{n'}$ be the samples from $p'$. 
Then to check 
resilience we need to bound $\|\frac{1}{|T|} \sum_{i \in T} (x_i - \mu)\|$ for all 
sets $T$ of size at least $(1-\epsilon)n'$. 
We will first use the covering $v_1, \ldots, v_M$ to obtain 
\begin{equation}
\Big\|\frac{1}{|T|} \sum_{i \in T} (x_i - \mu)\Big\| \leq 2 \max_{j=1}^M \frac{1}{|T|} \sum_{i \in T} \langle x_i - \mu, v_j \rangle.
\label{eq:vj-bound}
\end{equation}
The idea will be to analyze the sum over $\langle x_i - \mu, v_j \rangle$ for a fixed 
$v_j$ and then union bound over the $M$ possibilities. For a fixed $v_j$, we will split 
the sum into two components: those with small magnitude (roughly $\sigma/\epsilon$) and those 
with large magnitude (between $\sigma/\epsilon$ and $B$). 
We can then bound the former with Hoeffding's inequality, and using resilience we will 
be able to upper-bound the second moment of the latter, after which we can use 
Bernstein's inequality.

More formally, let $\tau = \frac{1-\epsilon}{\epsilon/4}\sigma$ and define
\begin{align}
y_i &= \langle x_i - \mu, v_j\rangle\bI[|\langle x_i - \mu, v_j \rangle| < \tau], \\
z_i &= \langle x_i - \mu, v_j\rangle\bI[|\langle x_i - \mu, v_j \rangle| \geq \tau].
\end{align}
Clearly $y_i + z_i = \langle x_i - \mu, v_j \rangle$. Also, we have 
$|y_i| \leq \tau$ almost surely, and $|z_i| \leq B$ almost surely (because $x_i \sim p'$ 
and hence $\langle x_i - \mu, v_j \rangle \leq \|x_i - \mu\| \leq B$).
The threshold $\tau$ is chosen so that $z_i$ is non-zero with probability at most 
$\epsilon/2$ under $p$ (see Lemma~\ref{lem:tail-bound}).

Now, for any set $T$ of size at least $(1-\epsilon)n'$, we have
\begin{align}
\frac{1}{|T|} \sum_{i \in T} \langle x_i - \mu, v_j \rangle
 &= \frac{1}{|T|} \sum_{i \in T} y_i + z_i \\
 &\leq \Big|\frac{1}{|T|} \sum_{i \in T} y_i\Big| + \frac{1}{|T|} \sum_{i \in T} |z_i| \\
 &\leq \Big|\frac{1}{|T|} \sum_{i=1}^{n'} y_i\Big| + \Big|\frac{1}{|T|} \sum_{i \not\in T} y_i\Big| + \frac{1}{|T|} \sum_{i=1}^{n'} |z_i| \\
 &\leq \frac{1}{1-\epsilon}\Big|\frac{1}{n'}\sum_{i=1}^{n'} y_i\Big| + \frac{\epsilon}{1-\epsilon}\tau + \frac{1}{(1-\epsilon)n'} \sum_{i=1}^{n'} |z_i|.
\label{eq:y-z-bound}
\end{align}
The last step uses the fact that $|y_i| \leq \tau$ for all $i$.
It thus suffices to bound $|\frac{1}{n'} \sum_{i=1}^{n'} y_i|$ as well as 
$\frac{1}{n'} \sum_{i=1}^{n'} |z_i|$.

For the $y_i$ term, note that by resilience $\|\bE[y_i]\| \leq \sigma$ (since $y_i$
is sampled from $p$ conditioned on $|\langle x_i - \mu, v_j \rangle| < \tau$ and 
$\|x_i - \mu\| \leq B$, which each occur with probabiliy at least $1-\epsilon/2$).
Then by Hoeffding's inequality, 
$|\frac{1}{n'} \sum_{i=1}^{n'} y_i| = \oo(\sigma + \tau\sqrt{\log(2/\delta)/n'})$ 
with probability $1-\delta$.

For the $z_i$ term, we note that $\bE[|z_i|] = \bE[\max(z_i,0)] + \bE[\max(-z_i,0)]$. 
Let $\tau'$ be the $\epsilon$-quantile of $\langle x_i - \mu, v_j \rangle$ under $p$, 
which by Lemma~\ref{lem:tail-bound} is at most $\tau$. Then we have
\begin{align}
\bE_{p}[\max(z_i,0)] &= \bE_{p}[\langle x_i - \mu, v_j \rangle\bI[\langle x_i - \mu, v_j \rangle \geq \tau]] \\
 &\leq \bE_{p}[\langle x_i - \mu, v_j \rangle\bI[\langle x_i - \mu, v_j \rangle \geq \tau']] \\
 &\stackrel{(i)}{\leq} \epsilon \cdot \frac{1-\epsilon}{\epsilon} \sigma = (1-\epsilon)\sigma,
\end{align}
where (i) is again Lemma~\ref{lem:tail-bound}.
Then we have $\bE_{p'}[\max(z_i,0)] \leq \frac{1}{1-\epsilon}\bE_{p}[\max(z_i,0)] \leq \sigma$, 
and hence $\bE_{p'}[|z_i|] \leq 2\sigma$ (as $\bE[\max(-z_i,0)] \leq \sigma$ by the same
argument as above). 
Since $|z_i| \leq B$, we then have $\bE[|z_i|^2] \leq 2B\sigma$.

Therefore, by 
Bernstein's inequality, with probability $1-\delta$ we have 
\begin{equation}
\frac{1}{n'} \sum_{i=1}^{n'} |z_i| \leq \oo\Big(\sigma + \sqrt{\frac{\sigma B \log(2/\delta)}{n'}} + \frac{B\log(2/\delta)}{n'}\Big) = \oo\Big(\sigma + \frac{B\log(2/\delta)}{n'}\Big).
\end{equation}
Taking a union bound over the $v_j$ for both $y$ and $z$, 
and plugging back into \eqref{eq:y-z-bound}, we get that 
$\frac{1}{|T|} \sum_{i \in T} \langle x_i - \mu, v_j \rangle \leq \oo\big(\sigma + \frac{\sigma}{\epsilon} \sqrt{\frac{\log(2M/\delta)}{n}} + \frac{B\log(2M/\delta)}{n}\big)$ for all $T$ and $v_j$ with probability $1-\delta$.
Plugging back into \eqref{eq:vj-bound}, we get that 
$\|\frac{1}{|T|} \sum_{i \in T} (x_i - \mu)\| \leq \oo\big(\sigma + \frac{\sigma}{\epsilon} \sqrt{\frac{\log(2M/\delta)}{n}} + \frac{B\log(2M/\delta)}{n}\big)$, 
as was to be shown.
\end{proof}

\subsection{Concentration Under Bounded Covariance}
\label{sec:concentration-cov}

In this section we state a stronger but more restrictive finite-sample bound 
giving conditions under which samples have bounded variance. 
It is a straightforward extension of Proposition B.1 of \citet{charikar2017learning}, 
so we defer the proof to Section~\ref{sec:concentration-cov-proof}.
\begin{proposition}
\label{prop:concentration-cov}
Suppose that a distribution $p$ has bounded variance in a 
norm $\|\cdot\|_*$: $\bE_{x \sim p}[\langle x-\mu, v \rangle^2] \leq \sigma^2 \|v\|_*^2$ 
for all $v \in \bR^d$. Then, given $n$ samples $x_1, \ldots, x_n \sim p$, 
with probability $1 - \exp(-\epsilon^2 n/16)$ there is a subset $T$ of 
$(1-\epsilon)n$ of the points such that 
\begin{equation}
\frac{1}{|T|} \sum_{i \in T} \langle x_i - \mu, v \rangle^2 \leq (\sigma')^2 \|v\|_*^2 \text{ for all } v \in \bR^d,
\end{equation}
where $(\sigma')^2 = \frac{4\sigma^2}{\epsilon} \p{1 + \frac{d}{(1-\epsilon)n}}$.
\end{proposition}
Proposition~\ref{prop:concentration-cov} says that whenever a distribution $p$ on $\bR^d$ 
has bounded variance, if $n \geq d$ samples $x_i$ are drawn from $p$ then some large subset of 
the samples will have bounded variance as well.

%% file: info-theory-applications.tex
\section{Information-Theoretic Applications}
\label{sec:info-theory-applications}

Using Proposition~\ref{prop:resilience} together with our finite-sample bounds, 
we establish information-theoretic robust recovery results in several cases of interest.

\subsection{Distribution Learning}
\label{sec:dist-learning}

First we consider the case where we are given $k$-tuples of independent samples 
from a distribution $\pi$ on $\{1,\ldots,m\}$. By taking the empirical distribution 
of the $k$ samples, we get an element of the simplex $\Delta_m$; for instance if 
$m = 5$ and $k = 3$ and a given triple of samples is $(2, 4, 2)$, then we produce 
the vector $(0, \frac{2}{3}, 0, \frac{1}{3}, 0) \in \Delta_5$. 
For general $\pi$ and $k$, we will denote the resulting distribution over $\Delta_m$ as 
$F(\pi, k)$, and think 
of both $\pi$ and samples $x \sim F(\pi, k)$ as elements of $\bR^m$.

We first establish finite-sample concentration for $F(\pi, k)$:
\begin{proposition}
\label{prop:dist-learning}
Let $p = F(\pi, k)$, where $\pi$ is a distribution on $\{1,\ldots,m\}$.
Then:
\begin{itemize}
\item Given $m$ samples $x_1, \ldots, x_m$ from $p$, 
      with probability $1-\exp(-\Omega(m))$ there is a subset 
      $T$ of $\frac{3m}{4}$ of the samples such that $\frac{1}{|T|} \sum_{i \in T} \langle x_i - \pi, v \rangle^2 \leq \oo\p{\frac{1}{k}}\|v\|_{\infty}^2$ for all $v \in \bR^d$.
      
\item Given $\frac{m\sqrt{k}}{\epsilon} + \frac{m}{\epsilon^2}$ 
      samples from $p$, with probability 
      $1 - \exp(-\Omega(m))$ there is a subset of $(1-\epsilon)m$ of the samples 
      that is $(\sigma, \epsilon)$-resilient in the $\ell_1$-norm, 
      with $\sigma = \oo(\epsilon \sqrt{\log(1/\epsilon)/k})$.
\end{itemize}
\end{proposition}
As a corollary, we immediately get robust recovery bounds for $\pi$. For instance, 
we have:
\begin{corollary}
\label{cor:dist-learning}
Suppose $\epsilon > 0$ is sufficiently small. Then, given 
$n = \Omega(\frac{m\sqrt{k}}{\epsilon} + \frac{m}{\epsilon^2})$ 
samples $x_1, \ldots, x_n$, $(1-\epsilon)n$ of which are drawn from $\pi^k$, 
with probability $1-\exp(-\Omega(m))$ it is information-theoretically possible to recover $\pi$ 
with total variational error $\oo(\epsilon \sqrt{\log(1/\epsilon)/k})$.
\end{corollary}
We omit the proof of Corollary~\ref{cor:dist-learning}, as it follows straightforwardly 
from Proposition~\ref{prop:dist-learning} and Proposition~\ref{prop:resilience}.
%the only non-trivial fact needed 
%is that $(\sigma,\epsilon)$-resilience implies $(2C\sigma, C\epsilon)$-resilience provided 
%$C \geq 1$, $C\epsilon \leq \frac{1}{2}$. %, which is a consequence of Lemma~\ref{lem:reverse}.
%
%We now prove Proposition~\ref{prop:dist-learning}
\begin{prooff}{Proposition~\ref{prop:dist-learning}}
Consider a sample $y \sim \pi$, interpreted as an indicator vector in $\{0,1\}^m$. 
First, we observe that for any vector $v \in \{-1,+1\}^m$, 
$\bE[\langle y, v \rangle^2] = \sum_{j=1}^m \pi_j v_j^2 = 1$, and so 
certainly $\bE[\langle y - \pi, v \rangle^2] \leq 1$. Therefore, if $x$ is an average of 
$k$ independent samples from $\pi$, $\bE[\langle x - \pi, v \rangle^2] \leq 1/k$ for all 
$v$ with $\|v\|_{\infty} \leq 1$. Invoking Proposition~\ref{prop:concentration-cov} then 
yields the first part of Proposition~\ref{prop:dist-learning}.

We next analyze the moment generating function $\bE[\exp(\lambda^{\top}(y-\pi))]$ in order 
to get a tail bound on $y-\pi$. Provided $\|\lambda\|_{\infty} \leq 1$, 
the moment generating function is bounded as 
\begin{align}
\bE[\exp(\lambda^{\top}(y-\pi))]
 &= \exp(-\lambda^{\top}\pi) \sum_{j=1}^m \pi_j \exp(\lambda_j) \\
 &\leq \exp(-\lambda^{\top}\pi) \sum_{j=1}^m \pi_j (1 + \lambda_j + \lambda_j^2) \\
 &\leq \exp(-\lambda^{\top}\pi) \exp(\sum_{j=1}^m \pi_j (\lambda_j + \lambda_j^2)) 
 = \exp(\sum_{j=1}^m \pi_j \lambda_j^2).
\end{align}
In particular, for any sign vector $v$ and $c \in [0,1]$ we have 
$\bE[\exp(cv^{\top}(y-\pi))] \leq \exp(c^2 \sum_{j=1}^m \pi_j) = \exp(c^2)$.
For an average $x$ of $k$ independent samples from $\pi$, 
%by multiplicativity of the moment generating functions, we have 
we therefore have 
$\bE[\exp(cv^{\top}(x-\pi))] \leq \exp(c^2/k)$ for $c \in [0,k]$. 
But note that $v^{\top}(x-\pi) \leq 2$, and so $\bE[\exp(cv^{\top}(x-\pi))] \leq \exp(2c^2/k)$ 
for all $c \geq k$ as well. Hence 
$\bE[\exp(cv^{\top}(x-\pi))] \leq \exp(2c^2/k)$ for all $c \geq 0$.

Now, let $E$ be any event of probability $\epsilon$. We have
\begin{align}
\bE[v^{\top}(x-\pi) \mid E] 
 &\leq \frac{1}{c}\log(\bE[\exp(cv^{\top}(x-\pi)) \mid E]) \\
 &\leq \frac{1}{c}\log(\frac{1}{\epsilon}\bE[\exp(cv^{\top}(x-\pi))]) \\
 &\leq \frac{1}{c}\log(\frac{1}{\epsilon}\exp(2c^2/k)) \\
 &= \frac{\log(1/\epsilon) + 2c^2/k}{c}.
\end{align}
Optimizing $c$ yields 
$\bE[v^{\top}(x-\pi) \mid E] \leq \sqrt{8\log(1/\epsilon)/k}$, 
and so (by Lemma~\ref{lem:reverse}) $p$ is $(\sigma,\epsilon)$-resilient around its mean in the $\ell_1$-norm, 
with $\sigma = \frac{\epsilon}{1-\epsilon}\sqrt{8\log(1/\epsilon)/k} = \oo(\epsilon\sqrt{\log(1/\epsilon)/k})$.

The second part of Proposition~\ref{prop:dist-learning} now follows 
immediately upon invoking Proposition~\ref{prop:concentration-resilient}, 
with $M = 2^d$ (for the $2^d$ sign vectors), 
$B = 2$ and $\sigma = \oo(\epsilon \sqrt{\log(1/\epsilon)/k})$.
\end{prooff} 

\subsection{Bounded $k$th Moments}

We next give results for distributions with bounded $k$th moments.
\begin{proposition}
Suppose that a distribution $p$ on $\bR^d$ has bounded 
$k$th moments: $\bE[|\langle x - \mu, v \rangle|^k]^{1/k} \leq \sigma_k \|v\|_2$ 
for all $v \in \bR^d$ and some $k \geq 2$. 
Then, given $n \geq \frac{d^{1.5}}{\epsilon} + \frac{d}{\epsilon^2}$ samples 
$x_1, \ldots, x_n \sim p$, with probability $1 - \exp(-\Omega(d))$ 
there is a set $T$ of $(1-\epsilon)n$ of the $x_i$ that is 
$(\oo(\sigma_k \epsilon^{1-1/k}),\epsilon)$-resilient in the $\ell_2$-norm.
\end{proposition}
\begin{proof}
We will invoke Proposition~\ref{prop:concentration-resilient}. 
Let $U$ be the uniform distribution on the unit sphere. First note that 
\begin{align}
\bE_{x}[\|x - \mu\|_2^k]
 &= \bE_{x}[(d\bE_{v \sim U}[|\langle x - \mu,v \rangle|^2])^{k/2}] \\
 &\leq d^{k/2} \bE_{x,v}[|\langle x - \mu, v \rangle|^k] \\
 &\leq d^{k/2} \sigma_k^k.
\end{align}
Therefore, by Chebyshev's inequality, $\bP[\|x - \mu\|_2 \geq (\epsilon/2)^{-1/k} \sigma_k\sqrt{d}] \leq \epsilon/2$. We can therefore take $B = \oo(\epsilon^{-1/k}\sigma_k\sqrt{d})$ 
in Proposition~\ref{prop:concentration-resilient}. Similarly, we can take 
$\sigma = \oo(\epsilon^{1-1/k}\sigma_k)$ in Proposition~\ref{prop:concentration-resilient}, 
and there is a covering of the $\ell_2$ ball with $9^d$ elements, so 
$\log(M) = \oo(d)$. Together, these imply that 
with probability $1 - \exp(-d)$ there is a set 
$T$ of size $(1-\epsilon)n$ that is $(\oo(\sigma_k\epsilon^{1-1/k}), \epsilon)$-resilient, 
as was to be shown.
\end{proof}
As before, we obtain the following immediate corollary:
\begin{corollary}
Let $p$ be a distribution on $\bR^d$ with $k$th moments bounded by $\sigma_k$, for $k \geq 2$, 
and suppose that $\epsilon > 0$ is sufficiently small. Then, given 
$n = \Omega(\frac{d^{1.5}}{\epsilon} + \frac{d}{\epsilon^2})$ samples $x_1, \ldots, x_n$, $(1-\epsilon)n$ of which 
are drawn from $p$, with probability $1 - \exp(-\Omega(d))$ it is possible to 
recover $\hat{\mu}$ within $\ell_2$-distance 
$\oo(\epsilon^{1-1/k}\sigma_k)$ of the mean of $p$.
\end{corollary}

\subsection{Stochastic Block Model}
\label{sec:sbm-app}

Our final set of information-theoretic results is for the semi-random stochastic block 
model from \citet{charikar2017learning}. 
In this model, we consider a graph $G$ on $n$ vertices, with an unknown set $S$ of 
$\alpha n$ ``good'' vertices. For simplicity we assume the graph is a directed graph. 
For $i, j \in S$, $i$ is connected to $j$ with probability $\frac{a}{n}$, and for 
$i \in S$, $j \not\in S$, $i$ is connected to $j$ with probability $\frac{b}{n}$, 
where $b < a$. For $i \not\in S$ the edges are allowed to be arbitrary.
If we let $A \in \{0,1\}^{n \times n}$ be the adjacency matrix of $G$, then the 
rows in $S$ are jointly independent samples from a distribution $p$ on $\{0,1\}^n$. 
The mean of this distribution is $\frac{a}{n}$ for the coordinates lying in $S$, and 
$\frac{b}{n}$ for the coordinates outside of $S$. 
We will apply our robust mean estimation results to this distribution, and use this 
to robustly recover the set $S$ itself. We let $x_1, \ldots, x_{\alpha n}$ denote the 
$\alpha n$ samples from $p$ corresponding to the elements of $S$.

\begin{lemma}
\label{lem:sbm-res}
\label{lem:sbm-resilience}
Suppose that $x_1,\ldots, x_{\alpha n}$ are drawn from a stochastic block model 
with parameters $(\alpha, a, b)$. Take the norm 
$\|x\| = \max_{|J| = \alpha n} \|x_J\|_1$, which is the maximum $\ell_1$-norm over any 
$\alpha n$ coordinates of $x$. Define $\mu_j = \frac{a}{n}$ if $j \in S$ and $\frac{b}{n}$ 
if $j\not\in S$. Then, with probability $1-\exp(-\Omega(\alpha n))$, the 
$x_i$ are $(\sigma, \alpha/2)$-resilient under $\|\cdot\|$ around $\mu$ with parameter
\begin{equation}
\sigma = \oo\p{\alpha \sqrt{a\log(2/\alpha)} + \log(2/\alpha)}.
\end{equation}
\end{lemma}

\begin{proof}
Note that we can express $\|x\|$ as $\max_{v \in \sV} \langle x, v \rangle$, 
where $\sV$ is the set of $\alpha n$-sparse $\{0,+1,-1\}$ vectors; in particular, 
$|\sV| = \binom{n}{\alpha n} 2^{\alpha n}$.
By Lemma~\ref{lem:reverse} and the definition of resilience, $\sigma$ is equal to
\begin{equation}
\frac{\alpha/2}{1-\alpha/2} \max_{T \subseteq \{1,\ldots,\alpha n\}, |T| = \frac{1}{2}\alpha^2 n} \max_{v \in \sV}
\Big\langle \frac{1}{|T|} \sum_{i \in T} x_i - \mu, v \Big\rangle.
\end{equation}
%By taking a covering $v^{(1)},\ldots,v^{(M)}$ of the $\alpha n$-dimensional sphere with 
%$M = 9^{\alpha d}$ elements, we can further bound $\sigma$ by 
%\begin{equation}
%\sigma \leq 2\max_{R \subseteq \{1,\ldots,\alpha n\}, |R| = \frac{1}{2}\alpha^2 n} \max_{T \subseteq \{1,\ldots, n\}, |T| = \alpha n} \max_{j=1}^M \left\langle \frac{1}{|R|} \sum_{i \in R} (x_i)_T - \mu_T, v^{(j)} \right\rangle.
%\end{equation}
We will union bound over all $T$ and $v$. 
For a fixed $T$ and $v$, the inner expression is equal to 
$\frac{2}{\alpha^2 n} \sum_{i \in T} \sum_{j=1}^n v_j (x_{ij} - \mu_j)$.
Note that all of the $v_j (x_{ij} - \mu_j)$ are independent, zero-mean random variables 
with variance at most $\frac{a}{n}v_j^2$ and are bounded in $[-1,1]$. 
By Bernstein's inequality, we have 
\begin{align}
\bP\left[\sum_{i \in T} \sum_{j=1}^n v_j (x_{ij} - \mu_j) \geq t\right]
 &\leq \exp\p{-\frac{1}{2}\frac{t^2}{\sum_{i, j} \frac{a}{n} v_j^2  + \frac{1}{3}t}} \\
 &= \exp\p{-\frac{t^2}{\alpha^3 a n  + \frac{2}{3}t}},
\label{eq:failure-prob}
\end{align}
%and hence with probability $1-\delta$ we have
%\begin{align}
%\frac{2}{\alpha^2 n} \sum_{i \in R, k \in T} v^{(j)}_k (x_{ik} - \mu_k) 
% &\leq \frac{2}{\alpha^2 n} \p{\sqrt{\alpha^2 a \log(1/\delta)} + \frac{3}{2}\log(1/\delta)} \\
% &\leq \sqrt{\frac{4a\log(1/\delta)}{\alpha^2 n^2}} + \frac{3\log(1/\delta)}{\alpha^2 n}.
%\end{align}
Now, if we want our overall union bound to hold with probability $1-\delta$, we need to 
set the term in \eqref{eq:failure-prob} to be at most 
$\delta / \left[\binom{\alpha n}{\alpha^2 n/2}\binom{n}{\alpha n}2^{\alpha n}\right]$, 
so $\frac{t^2}{\alpha^3 an + \frac{2}{3}t} = \log(1/\delta) + \oo\p{\alpha n\log(2/\alpha)} = \oo\p{\alpha n \log(2\alpha)}$ (since we will take $\delta = \exp(-\Theta(\alpha n))$). 
Hence we can take 
%\begin{equation}
%t = \oo\p{\sqrt{\alpha^3 an \log(1/\delta)} + \sqrt{\alpha^4 an^2\log(2/\alpha)} + \log(1/\delta) + \alpha n \log(2/\alpha)}.
$t = \oo\p{\sqrt{\alpha^4 an^2\log(2/\alpha)} + \alpha n \log(2/\alpha)}$.
%\end{equation}
Dividing through by $\frac{1}{2}\alpha^2 n$ and multiplying by $\frac{\alpha/2}{1-\alpha/2}$, 
we get
\begin{equation}
\sigma = \oo\p{\alpha \sqrt{a\log(2/\alpha)} + \log(2/\alpha)},
\end{equation}
as was to be shown.
\end{proof}

Using Lemma~\ref{lem:sbm-resilience}, we obtain our result on robust recovery of the 
set $S$:
\begin{corollary}
Under the semi-random stochastic block model with parameters 
$\alpha$, $a$, and $b$, 
it is information-theoretically possible to obtain a set $\hat{S}$ satisfying
\begin{equation}
\frac{1}{\alpha n}|\hat{S} \triangle S| = \oo\p{\sqrt{\frac{a\log(2/\alpha)}{\alpha^2(a-b)^2}}}.
\end{equation}
with probability $\Omega(\alpha)$.
\end{corollary}
\begin{proof}
By Lemma~\ref{lem:sbm-res} and Proposition~\ref{prop:resilience}, 
we can recover a $\hat{\mu}$ that with probability $\Omega(\alpha)$ satisfies 
\begin{equation}
\label{eq:sbm-res-1}
%\|\hat{\mu} - \mu\| = \oo\p{\sqrt{\frac{a\log(1/\delta)}{\alpha n}} + \sqrt{a\log(2/\alpha)} + \frac{\log(1/\delta)}{\alpha^2 n} + \frac{\log(2/\alpha)}{\alpha}}.
\|\hat{\mu} - \mu\| = \oo\p{\sqrt{a\log(2/\alpha)} + \frac{\log(2/\alpha)}{\alpha}}.
\end{equation}
Note that $\mu_j = \frac{a}{n}$ if $j \in S$ and $\frac{b}{n}$ if $j \not\in S$, where 
$b < a$. We will accordingly define $\hat{S}$ to be the set of coordinates 
$j$ such that $\hat{\mu}_j \geq \frac{a+b}{2n}$. We then have that 
$\|\hat{\mu} - \mu\| = \Omega(\frac{a-b}{n}\min(|\hat{S} \triangle S|, \alpha n))$, and hence 
$\frac{1}{\alpha n}|\hat{S} \triangle S| = \oo(\frac{1}{\alpha (a-b)}\|\hat{\mu} - \mu\|)$ 
whenever the right-hand-side is at most $1$. Using \eqref{eq:sbm-res-1}, we have that 
\begin{align}
\frac{1}{\alpha n}|\hat{S} \triangle S| 
 &= \oo\p{\frac{1}{\alpha (a-b)}\p{\sqrt{a\log(2/\alpha)} + \frac{\log(2/\alpha)}{\alpha}}} \\
 &= \oo\p{\sqrt{\frac{a\log(2/\alpha)}{\alpha^2(a-b)^2}} + \frac{\log(2/\alpha)}{\alpha^2 (a-b)}} 
 \leq \oo\p{\sqrt{\frac{a\log(2/\alpha)}{\alpha^2(a-b)^2}} + \frac{a\log(2/\alpha)}{\alpha^2 (a-b)^2}}
\label{eq:sbm-bound}
\end{align}
Here the last step simply multiplies the second term by $\frac{a}{a-b}$, which is at least $1$.
Note that the first term in \eqref{eq:sbm-bound} dominates whenever the bound is meaningful, 
%Since the bound on $\frac{1}{\alpha n}|\hat{S} \triangle S|$ 
%is only meaningful when it is at most $1$, we can assume that the first term dominates, which 
which yields the desired result.
\end{proof}

%Suppose that we have $\alpha n$ points $x_1, \ldots, x_{\alpha n}$ 
%drawn from a stochastic block model, 
%meaning that $x_i \in [0,1]^n$, and $x_{ij} \sim \Ber(a/n)$ if $j \in S$ and 
%$x_{ij} \sim \Ber(b/n)$ if $j \not\in S$, where $S$ also has size $\alpha n$.
%
%We will define a norm $\|v\| = \max(\|v\|_{\infty}, \frac{1}{\alpha n} \|v\|_1)$. 
%The unit ball in this norm is the convex hull of $\{0,+1,-1\}$ vectors with 
%$\alpha n$ non-zero coordinates. Therefore, we can take $M = 2^{\alpha n} \binom{n}{\alpha n}$ 
%in Proposition~\ref{prop:concentration}, and $\log M = \oo(\alpha n \log(2/\alpha))$. 
%
%Let $\mu_j = a/n$ if $j \in S$ and $b/n$ if $j\not\in S$. 
%For any vector $v$, we have the cumulant function inequality
%\begin{align}
%\bE_x\Big[\exp\big(\lambda \sum_{j=1}^n v_j(x_j - \mu_j)\big)\Big]
% &\leq \prod_{j=1}^n (1 + (a/n)(e^{\lambda v_j} - 1)
%\end{align}

%% file: algo-applications.tex
\section{Algorithmic Applications}
\label{sec:algo-applications}

We end by presenting applications of our algorithmic result Theorem~\ref{thm:alg-intro}.
First, we show that Theorem~\ref{thm:alg-intro} implies a deterministic recovery 
result for sets that are resilient in the $\ell_p$-norm, for any $p \in (1,2]$:
\begin{corollary}
\label{cor:lp}
Suppose that a set of points $x_1, \ldots, x_n$ contains a set $S$ of size $\alpha n$ that
is $(\sigma,\frac{1}{2})$-resilient around its mean $\mu$ in $\ell_p$-norm, with $p \in (1,2]$.
Then there is an efficient algorithm whose output $\hat{\mu}$
satisfies $\|\mu - \hat{\mu}\|_p = \oo\Big({\frac{\sigma}{\alpha(p-1)}}\Big)$ with 
probability $\Omega(\alpha)$.

Moreover, if $\alpha = 1-\epsilon \geq 0.87$ and $S$ is 
$(\sigma\sqrt{\delta}, \delta)$-resilient 
for $\delta \in [\frac{\epsilon}{2}, \frac{1}{2}]$, the same algorithm has
$\|\mu - \hat{\mu}\|_p = \oot\Big({\sigma\sqrt{\frac{\epsilon}{p-1}}}\Big)$
with probability $1$. In particular, this holds if $S$ has variance bounded by 
$\sigma^2$.
\end{corollary}
\begin{proof}
First, we observe that the $\ell_p$-norm is $(p-1)$-strongly convex for $p \in (1,2]$ 
(c.f. Lemma 17 of \citet{shalev07online}). Therefore, if $S$ has the assumed level of 
resilience, then we can find a core $S_0$ of size $\frac{\alpha}{2}$ with second moments 
bounded by $\oo(\frac{\sigma^2}{p-1})$ by Proposition~\ref{prop:powering-up-lp}. 
Moreover, if $\alpha = 1-\epsilon \geq 0.87$, then we can find a core of size 
$(1-\epsilon)^2 n \geq \frac{3}{4}n$ by Proposition~\ref{prop:powering-up-lp-eps}.
In this latter case, the core will have second moments bounded by $\oo(\frac{\sigma^2 \log(1/\epsilon)}{p-1})$.

In each case, we can apply Theorem~\ref{thm:alg-intro} to this bounded-variance core.
We need to check that Assumption~\ref{ass:opt} holds. This is due to the 
following result, which follows from Theorem 3 of \citet{nesterov1998semidefinite}:
\begin{theorem}[\citeauthor{nesterov1998semidefinite}]
\label{thm:grothendieck}
Let $\|\cdot\|_q$ denote the $\ell_q$-norm for $q \geq 2$. 
For a matrix $A \in \bR^{d \times d}$, let $f(A) = \max_{\|v\|_q \leq 1} v^{\top}Av$, 
and let $g(A) = \max_{Y \succeq 0, \|\diag(Y)\|_{q/2} \leq 1} \langle A, Y \rangle$. 
Then $f(A) \leq g(A) \leq \frac{\pi}{2} f(A)$.
\end{theorem}
In particular, for the $\ell_p$-norm with $p \in [1,2]$, the dual norm is the 
$\ell_q$-norm with $q = \frac{p}{p-1} \in [2,\infty]$. Therefore, Theorem~\ref{thm:grothendieck} 
asserts that Assumption~\ref{ass:opt} holds for $\sP = \{Y \mid Y \succeq 0, \|\diag(Y)\|_{q/2} \leq 1\}$ and $\kappa = \frac{\pi}{2} = \oo(1)$.

Now, applying Theorem~\ref{thm:alg-intro} for general $\alpha$, 
this yields a recovery guarantee of 
$\|\hat{\mu} - \mu\|_p = \oo\big(\frac{\sigma}{\alpha (p-1)}\big)$ with probability 
$\Omega(\alpha)$, while for $\alpha = 1-\epsilon \geq 0.87$ this yields a recovery 
guarantee of $\|\hat{\mu} - \mu\|_p = \oo\big(\frac{\sigma}{\sqrt{p-1}} \sqrt{\epsilon \log(1/\epsilon)}\big)$ with probability $1$, as claimed.
\end{proof}

We next specialize Corollary~\ref{cor:lp} to the distribution learning case and 
provide finite-sample bounds.

\begin{corollary}
\label{cor:dist-learning-alg}
For the distribution learning problem from Section~\ref{sec:dist-learning}, 
suppose that we are given $n$ samples, of which $\alpha n$ are independent $k$-tuples from a 
distribution $\pi$ on $\{1,\ldots,m\}$, and the remaining samples are arbitrary. Then:
\begin{itemize}
\item If $\alpha n \geq m$, with probability $1 - \exp(-\Omega(m))$ there is a randomized 
      algorithm outputting a $\hat{\pi}$ satisfying 
      $\|\hat{\pi} - \pi\|_{TV} = \tilde{\oo}(\frac{1}{\alpha\sqrt{k}})$ with 
      probability $\Omega(\alpha)$.
\item If $\alpha = 1-\epsilon$, where $\epsilon$ is sufficiently small, 
      and $n \geq \frac{m\sqrt{k}}{\epsilon} + \frac{m}{\epsilon^2}$, then with 
      probability $1 - \exp(-\Omega(m))$ there is a deterministic 
      algorithm outputting a $\hat{\pi}$ satisfying 
      $\|\hat{\pi} - \pi\|_{TV} = \tilde{\oo}(\sqrt{\epsilon/k})$. % with probability $1$.
\end{itemize}
In both of the above bounds, the $\tilde{\oo}$ notation hides polylog factors in both 
$\epsilon$ and $m$.
\end{corollary}
\begin{proof}
By Proposition~\ref{prop:dist-learning}, in the first case there will with high probability 
be a set of size $\frac{3}{4}\alpha n$ with bounded variance under the $\ell_1$-norm. 
Since the $\ell_1$-norm and $\ell_{1 + \frac{1}{\log(m)}}$-norm are within constant 
factors of each other, the set will also have bounded variance in the 
$\ell_{1 + \frac{1}{\log(m)}}$-norm, which is $\Omega(\frac{1}{\log(m)})$-strongly convex. 
We can then invoke 
Corollary~\ref{cor:lp} to obtain error $\oo(\frac{\sigma \log(m)}{\alpha})$.

In the second case, Proposition~\ref{prop:dist-learning} says that we can remove 
$\epsilon |S|/2$ points from $S$ to make the remaining set 
$(\oo(\epsilon \sqrt{\log(1/\epsilon)/k}), \epsilon/2)$-resilient. 
This implies that it is also $(\oo(\delta \sqrt{\log(1/\epsilon)/k}), \delta)$-resilient 
for all $\delta \in [\frac{\epsilon}{2}, \frac{1}{2}]$, and so by 
Proposition~\ref{prop:powering-up-lp-eps} it 
has a core of size $(1-\epsilon)(1-\epsilon/2)|S| = (1-\oo(\epsilon)) n$ 
with variance at most $\frac{\log(m)\log(1/\epsilon)}{k}$ (we again need to pass from 
$\ell_1$ to $\ell_{1 + \log(m)}$ to apply Proposition~\ref{prop:powering-up-lp-eps}, which 
is why we incur the $\log(m)$ factor). 

Applying Theorem~\ref{thm:alg-intro} as in the second part of Corollary~\ref{cor:lp} then 
yields the desired error bound of $\oo(\sqrt{\epsilon\log(1/\epsilon)\log(m)/k}) = \tilde{\oo}(\sqrt{\epsilon/k})$. %as was to be shown.
\end{proof}

%% file: counterexample.tex
\section{Counterexample to Analogues of Proposition~\ref{prop:powering-up-lp}}
\label{sec:counterexample}

In this section we show:
\begin{proposition}
\label{prop:counterexample}
Let $S = \{e_1, \ldots, e_n\}$, where the $e_i$ are the standard basis 
in $\bR^n$. Then:
\begin{enumerate}
\item $S$ is $(2n^{1/p-1}, \frac{1}{2})$-resilient around $0$ in $\ell_p$-norm, 
      for all $p \in [1,\infty]$.
\item Any subset $T$ of $\frac{n}{2}$ elements of $S$ has $\max_{\|v\|_q \leq 1} \frac{1}{|T|} \sum_{i \in T} |\langle e_i, v \rangle|^k = (n/2)^{\max(-1, k(\frac{1}{p}-1))}$. %/q)}$.
\end{enumerate}
\end{proposition}
Here we let $\|\cdot\|_q$ denote the dual norm to $\|\cdot\|_p$, i.e. $q$ is such that 
$\frac{1}{p} + \frac{1}{q} = 1$.

Proposition~\ref{prop:counterexample} implies in particular that for large $n$, an analog of 
Proposition~\ref{prop:powering-up-lp} can only hold if $k(1/p-1) \geq -1$, 
which implies that $k \leq \frac{p}{p-1}$. For $k = 2$, this implies $p \leq 2$ (so the norm 
is strongly convex), 
and for $p = 2$ this implies $k \leq 2$. We do not know whether an 
analog of Proposition~\ref{prop:powering-up-lp} holds for e.g. $k = 3$, $p = 1.5$.
%This implies that on the one hand $\frac{1}{p} - 1 \geq -\frac{1}{2}$, and hence 
%$p \leq 2$. On the other hand $\frac{1}{p} - 1 \geq \frac{k}{2}(\frac{1}{p}-1)$, 
%and hence $k \leq 2$. Therefore Proposition~\ref{prop:powering-up-lp} cannot hold 
%if $p > 2$ or $k > 2$, so that its current range of applicability is essentially optimal 
%for $\ell_p$-norms.

\begin{prooff}{Proposition~\ref{prop:counterexample}}
For the first part, note that the mean of any subset of $\frac{n}{2}$ elements of $S$ will 
have all coordinates lying in $[0, \frac{2}{n}]$. Therefore, the $\ell_p$-norm of the mean 
is at most $\frac{2}{n} \cdot n^{1/p} = 2n^{1/p-1}$, as claimed.

For the second part, assume without loss of generality that $T = \{e_1, \ldots, e_{n/2}\}$. 
%in which case the last $n/2$ coordinates of the maximizing $v$ will be zero.
The optimization then reduces to $\max_{\|v\|_q \leq 1} \frac{2}{n} \|v\|_k^k$, where 
$v \in \bR^{n/2}$. For the optimal $v$, it is clear that 
$\|v\|_k = 1$ if $k \geq q$, and $\|v\|_k = (n/2)^{\frac{1}{k} - \frac{1}{q}}$ if $k < q$, whence 
$\frac{2}{n} \|v\|_k^k = \frac{2}{n} \cdot (n/2)^{\max(0, 1-\frac{k}{q})} = (n/2)^{\max(-1, -\frac{k}{q})} = (n/2)^{\max(-1, -k(1-\frac{1}{p}))}$, as claimed.
\end{prooff}

%% file: resilience-properties.tex
\vskip -0.3in
\phantom{x}
\section{Basic Properties of Resilience}
\label{sec:resilience-properties}

Here we prove Lemmas~\ref{lem:tail-bound} and \ref{lem:reverse}.
Recall that the first states that resilience around the mean $\mu$ is equivalent to 
bounded tails, while the latter relates $\epsilon$-resilience to $(1-\epsilon)$-resilience.

\begin{prooff}{Lemma~\ref{lem:reverse}}
Note that for a distribution $p$, resilience around $\mu$ can be computed as 
$\sigmabest(\epsilon) = \max\{\|\bE[X \mid E] - \mu\| \mid E \text{ has probability } 1-\epsilon\}$.
When $\mu$ is the mean, then $\mu = \bE[X]$. We can then write 
\begin{align}
 &= \|\bE[X \mid E] - \bE[X]\| \\
 &= \|\bE[X \mid E] - \bP[E]\bE[X \mid E] - \bP[\neg E]\bE[X \mid \neg E]\| \\
 &= (1-\bP[E])\|\bE[X \mid E] - \bE[X \mid \neg E]\|.
 %&= \tfrac{1-\bP[E]}{\bP[E]} \|\bE[X \mid \neg E] - \bE[X]\|.
\label{eq:res-sym}
\end{align}
From \eqref{eq:res-sym}, it is clear that 
$\sigmabest(1-\epsilon) = \frac{1-\epsilon}{\epsilon}\sigmabest(\epsilon)$, 
as \eqref{eq:res-sym} is invariant under replacing $E$ with $\neg E$, except that 
$1-\bP[E]$ becomes $\bP[E]$.

More generally, if $\mu$ is not the mean then we still have 
$\|\mu - \bE[X]\| = \sigmabest(0) \leq \sigmabest(\epsilon)$ by resilience. Then if 
$E$ is an event with probability $\epsilon$, we have
\begin{align}
\|\bE[X \mid E] - \mu\| 
 &\leq \|\bE[X \mid E] - \bE[X]\| + \|\bE[X] - \mu\| \\
 &\leq \frac{1-\epsilon}{\epsilon}\|\bE[X \mid \neg E] - \bE[X]\| + \sigmabest(\epsilon) \\
 &\leq \frac{1-\epsilon}{\epsilon}(\|\bE[X \mid \neg E] - \mu\| + \|\mu - \bE[X]\|) + \sigmabest(\epsilon) \\
 &\leq \frac{1-\epsilon}{\epsilon}\cdot 2\sigmabest(\epsilon) + \sigmabest(\epsilon) = \frac{2-\epsilon}{\epsilon}\sigmabest(\epsilon).
\end{align}
Therefore, even when $\mu$ is not the mean we have 
$\sigmabest(1-\epsilon) \leq \frac{2-\epsilon}{\epsilon}\sigmabest(\epsilon)$, 
which completes the proof.
\end{prooff}

\begin{prooff}{Lemma~\ref{lem:tail-bound}}
We note that 
\begin{align}
\sigmabest(\epsilon) 
 &\stackrel{(i)}{=} \frac{1-\epsilon}{\epsilon}\sigmabest(1-\epsilon) \\
 &\stackrel{(ii)}{=} \frac{1-\epsilon}{\epsilon}\max_E\{\|\bE[X-\mu \mid E]\| \mid \bP[E] \geq \epsilon\} \\
 &\stackrel{(iii)}{=} \frac{1-\epsilon}{\epsilon}\max_{E,v}\{\langle \bE[X-\mu \mid E], v \rangle \mid \bP[E] \geq \epsilon, \|v\|_* \leq 1\} \\
 &\stackrel{(iv)}{=} \frac{1-\epsilon}{\epsilon}\max_{E,v}\{\bE[\langle X-\mu , v \rangle \mid E] \mid \bP[E] \geq \epsilon, \|v\|_* \leq 1\} \\
 &\stackrel{(v)}{=} \frac{1-\epsilon}{\epsilon}\max_{v}\{\bE[\langle X-\mu , v \rangle \mid \langle X-\mu, v \rangle \geq \tau_{\epsilon}(v)] \mid \|v\|_* \leq 1\},
\end{align}
as was to be shown.
Here (i) is Lemma~\ref{lem:reverse}, (ii) is the definition of resilience, and (iii) is 
the definition of the dual norm. Then (iv) uses linearity of expectation, while (v) 
explicitly maximizes $\bE[\langle X-\mu, v \rangle \mid E]$ over events $E$ with probability 
at least $\epsilon$.
\end{prooff}
%\begin{lemma}
%Suppose that 
%$\frac{1}{|S|} \sum_{i \in S} |\langle x_i - \mu, v \rangle|^2 \leq \sigma^2 \|v\|_*^2$ 
%for all $v \in \bR^d$, where $\mu$ is the empirical mean of the $x_i$. Then $S$ is 
%$(\sigma\sqrtt{\frac{\epsilon}{1-\epsilon}}, \epsilon)$-resilient around 
%$\mu$ for all $\epsilon$.
%\end{lemma}
%\begin{proof}
%Without loss of generality take $\mu = 0$ and denote the norm as $\|\cdot\|_{\psi}$. 
%The condition is equivalent to 
%$\|X\|_{2 \to \psi} \leq \sigma\sqrt{|S|}$. For any set $T$ define the vector 
%$w \in \bR^n$ by $w_i = \frac{\bI[i \in T]}{|T|} - \frac{1}{|S|}$. Then 
%$\|Xw\|_{\psi} = \|\mu_T - \mu\|_{\psi}$, where $\mu_T$ is the mean over $T$. 
%On the other hand, if $|T| = (1-\epsilon)n$ then 
%$\|Xw\|_{\psi} \leq \|X\|_{2 \to \psi}\|w\|_2 \leq \sigma\sqrt{|S|} \cdot \sqrt{|T| (\frac{1}{|T|} - \frac{1}{|S|})^2 + (|S|-|T|)\frac{1}{|S|^2}} = \sigma\sqrt{\frac{\epsilon}{1-\epsilon}}$, 
%which yields the desired result.
%\end{proof}

%% file: 1st-moment-lp-proof.tex
\section{Proof of Lemma~\ref{lem:1st-moment-lp}}
\label{sec:1st-moment-lp-proof}

For a fixed $v$, let $S_+ = \{i \mid \langle x_i - \mu, v \rangle > 0\}$, 
and define $S_-$ similarly. Either $S_+$ or $S_-$ must have size at least 
$\frac{1}{2}|S|$, without loss of generality assume it is $S_+$. Then we have
\begin{align}
\frac{1}{|S|} \sum_{i \in S} |\langle x_i - \mu, v \rangle| 
 &= \frac{1}{|S|} \bigg({ 2\sum_{i \in S_+} \langle x_i - \mu, v \rangle - \sum_{i \in S} \langle x_i - \mu, v \rangle}\bigg) \\
 &\leq \frac{1}{|S|} \bigg({2\|\sum_{i \in S_+} (x_i - \mu)\|\|v\|_* + \|\sum_{i \in S} (x_i - \mu)\|\|v\|_*}\bigg) \\
 &\leq \bigg({2\frac{|S_+|}{|S|} + 1}\bigg)\sigma \|v\|_* \leq 3\sigma \|v\|_*,
\end{align}
which completes the first part. In the other direction, for any set $T$, we have
\begin{align}
\Big\|\frac{1}{|T|} \sum_{i \in T} (x_i - \mu)\Big\| 
 &= \max_{\|v\|_* = 1} \frac{1}{|T|} \sum_{i \in T} \langle x_i - \mu, v \rangle  \\
 %\leq \max_{\|v\|_* = 1} \frac{1}{|T|} \sum_{i \in T} |\langle x_i - \mu, v \rangle| \\
 &\leq \max_{\|v\|_* = 1} \frac{1}{|T|} \sum_{i \in S} |\langle x_i - \mu, v \rangle| 
 \leq \frac{|S|}{|T|} \sigma \leq 2\sigma,
\end{align}
as was to be shown. \qed

%% file: minimax-proof.tex
\section{Proof of Lemma~\ref{lem:minimax}}
\label{sec:minimax-proof}

We start by taking a continuous relaxation of \eqref{eq:minimax-0}, asking for 
weights $c_i \in [0,1]$ rather than $\{0,1\}$:
\begin{equation}
\label{eq:minimax}
\min_{c \in [0,1]^n, \|c\|_1 \geq \frac{3n}{4}} \max_{\|v\|_* \leq 1} \frac{1}{n} \sum_{i=1}^n c_i |\langle x_i, v \rangle|^2.
\end{equation}
Note that we strengthened the inequality to $\|c\|_1 \geq \frac{3n}{4}$, whereas in 
\eqref{eq:minimax-0} it was $\|c\|_1 \geq \frac{n}{2}$.
Given any solution $c_{1:n}$ to \eqref{eq:minimax}, we can obtain a solution $c'$ to 
\eqref{eq:minimax-0} by letting $c_i' = \bI[c_i \geq \frac{1}{2}]$. 
Then $c_i' \in \{0,1\}$ and $\|c'\|_1 \geq \frac{n}{2}$. Moreover, $c_i' \leq 2c_i$, 
so $\frac{1}{n}\sum_{i=1}^n c_i' |\langle x_i, v \rangle|^2 \leq \frac{2}{n} \sum_{i=1}^n c_i |\langle x_i, v \rangle|^2$ for all $v$. Therefore, the value of \eqref{eq:minimax-0} is at 
most twice the value of \eqref{eq:minimax}.

Now, by the minimax theorem, we can swap the min and max in \eqref{eq:minimax} in 
exchange for replacing the single vector $v$ with a distribution over vectors $v_j$, 
thus obtaining that \eqref{eq:minimax} is equal to
\begin{equation}
\lim_{m \to \infty} \max_{\substack{\alpha_1 + \cdots + \alpha_m \leq 1 \\ \alpha \geq 0, \|v_j\|_* \leq 1}} \, \min_{\substack{c \in [0,1]^n \\ \|c\|_1 \geq \frac{3n}{4}}} \frac{1}{n} \sum_{i=1}^n c_i \sum_{j=1}^m \alpha_j |\langle x_i, v_j \rangle|^2.
\end{equation}
By letting $v_j' = \alpha_jv_j$, the above is equivalent to optimizing over 
$v_j$ satisfying $\sum_j \|v_j\|_*^2 \leq 1$:
\begin{equation}
\lim_{m \to \infty} \max_{\|v_1\|_*^2 + \cdots + \|v_m\|_*^2 \leq 1} \, \min_{\substack{c \in [0,1]^n \\ \|c\|_1 \geq \frac{3n}{4}}} \frac{1}{n} \sum_{i=1}^n c_i \sum_{j=1}^m |\langle x_i, v_j \rangle|^2.
\end{equation}
For any $v_1, \ldots, v_m$, we will find $c$ such that 
the above sum is bounded. Indeed, define $B(v_{1:m})$ to be 
$\frac{1}{n} \sum_{i=1}^n \sqrt{\sum_{j=1}^m |\langle x_i, v_j \rangle|^2}$. 
Then take 
$c_i = \bI[\sum_{j=1}^m |\langle x_i, v_j \rangle|^2 < 16B^2]$,
which has $\|c\|_1 \geq \frac{3n}{4}$ by Markov's inequality, 
and for which $\sum_{i} c_i \sum_j |\langle x_i, v_j \rangle|^2 \leq 4B(v_{1:m})^2$.

Therefore, the value of \eqref{eq:minimax} is bounded by 
$\max_{m, v_{1:m}} 4B(v_{1:m})^2$, 
and so the value of \eqref{eq:minimax-0} is bounded by 
$\max_{m, v_{1:m}} 8B(v_{1:m})^2$, which yields 
the desired result. \qed

%% file: powering-up-lp-eps-proof.tex
\section{Proof of Proposition~\ref{prop:powering-up-lp-eps}}
\label{sec:powering-up-lp-eps-proof}

%\begin{proof}[Proof of Proposition~\ref{prop:powering-up-lp-eps}]
The proof mirrors that of Proposition~\ref{prop:1st-2nd-lp}, but with more 
careful bookkeeping. Instead of \eqref{eq:minimax}, we consider the minimax problem
\begin{equation}
\label{eq:minimax-eps}
\min_{\substack{c \in [0,1]^n, \\ \|c\|_1 \geq (1-\epsilon/2)n}} \max_{\|v\|_* \leq 1} \frac{1}{n} \sum_{i=1}^n c_i |\langle x_i, v \rangle|^2.
\end{equation}
The only difference is that $\frac{3}{4}n$ has been replaced with $(1-\epsilon/2)n$ in 
the constraint on $\|c\|_1$. We then end up needing to bound
\begin{equation}
\max_{\sum_j \|v_j\|_*^2 \leq 1} \frac{1}{n} \sum_{k=1}^{(1-\frac{\epsilon}{2})n} \sum_{j=1}^m |\langle x_{i_k}, v_j \rangle|^2.
\end{equation}
Here we suppose that the indices $i$ are arranged in order $i_1, i_2, \ldots, i_n$ such that 
the output of the inner sum is monotonically increasing in $i_k$ 
(in other words, the outer sum over $k$ is over 
the $(1-\frac{\epsilon}{2})n$ smallest values of the inner sum, which corresponds to the 
optimal choice of $c$). Now we have
\begin{align}
\frac{1}{n} \sum_{k=1}^{(1-\frac{\epsilon}{2})n} \sum_{j=1}^m |\langle x_{i_k}, v_j \rangle|^2 
 &\stackrel{(i)}{\leq} \frac{1}{n} \sum_{k=1}^{(1-\frac{\epsilon}{2})n} \Bigg({\frac{1}{n-k+1} \sum_{l=k}^n \sqrt{\sum_{j=1}^m |\langle x_{i_l}, v_j \rangle|^2}}\Bigg)^2 \\
 &\stackrel{(ii)}{\leq} \frac{2}{n} \sum_{k=1}^{(1-\frac{\epsilon}{2})n} \Bigg({\frac{1}{n-k+1} \bE_{s}\Bigg[\sum_{l=k}^n \Big| \Big\langle x_{i_l}, \sum_{j=1}^m s_jv_j \Big\rangle\Big|\Bigg]}\Bigg)^2 \\
 &\stackrel{(iii)}{\leq} \frac{2}{n} \sum_{k=1}^{(1-\frac{\epsilon}{2})n} \Bigg({\frac{3n\min(\sigmabest(\frac{n-k+1}{n}),\sigmabest(\frac{1}{2}))}{n-k+1} \bE_{s}\Bigg[\Big\|\sum_{j=1}^m s_jv_j\Big\|_*\Bigg]}\Bigg)^2 \\
 &\leq \frac{18}{n\gamma} \sum_{k=1}^{(1-\frac{\epsilon}{2})n} \Bigg({\frac{\min(\sigmabest(\frac{n-k+1}{n}), \sigmabest(\frac{1}{2}))}{\frac{n-k+1}{n}}}\Bigg)^2.
\label{eq:pre-integral}
\end{align}
Here (i) bounds the $k$th smallest term by the average of the $n-k+1$ largest terms, 
(ii) is Khinchine's inequality, and (iii) is the following lemma which we prove later:
\begin{lemma}
\label{lem:1st-eps}
If $S$ is resilient around $\mu$ and $T \subseteq S$ has size at most $\epsilon |S|$, 
then $\sum_{i \in T} |\langle x_i - \mu, v \rangle| \leq 3|S|\|v\|_* \cdot \min(\sigmabest(\epsilon),\sigmabest(\frac{1}{2}))$.
\end{lemma}
Continuing, we can bound \eqref{eq:pre-integral} by 
\begin{align}
\frac{18}{\gamma} \int_{0}^{1-\frac{\epsilon}{2}} \bigg({\frac{\min(\sigmabest(1-u), \sigmabest(\frac{1}{2}))}{1-u - 1/n}}\bigg)^2 du
 &\leq \frac{18}{\gamma}\bigg({\frac{1}{\frac{\epsilon}{2} - \frac{1}{n}}}\bigg)^2 \int_{\epsilon/2}^{1} \bigg({\frac{\min(\sigmabest(u), \sigmabest(\frac{1}{2}))}{u}}\bigg)^2 du \\
 &\leq \oo(1) \cdot \frac{1}{\gamma} \cdot \bigg({ \int_{\epsilon/2}^{1/2} u^{-2} \sigmabest(u)^2 du + 2\sigmabest(\tfrac{1}{2})^2}\bigg) \\
 &= \oo\p{\frac{1}{\gamma} \cdot \p{\sigmastar(\epsilon)^2 + \sigmabest(\tfrac{1}{2})^2}}.
\end{align}
Since the $\sigmastar(\epsilon)$ term dominates, this completes the bound on 
\eqref{eq:minimax-eps}, from which the remainder of the proof follows 
identically to Proposition~\ref{prop:powering-up-lp}. \qed
%\end{proof}
%
We end this section by proving Lemma~\ref{lem:1st-eps}.
\begin{prooff}{Lemma~\ref{lem:1st-eps}}
Note that the $\sigmabest(\frac{1}{2})$ part of the bound was already established in 
Lemma~\ref{lem:1st-moment-lp}, so it suffices to show the $\sigmabest(\epsilon)$ part.
Let $\hat{\mu}$ be the mean of $S$. Note that we must 
have $\|\mu - \hat{\mu}\| \leq \sigmabest(\epsilon)$ by definition. 
We can thus replace $\mu$ with $\hat{\mu}$ in the statement of Lemma~\ref{lem:1st-eps} 
at the cost of changing the left-hand-side by at most 
$|T|\sigmabest(\epsilon)\|v\|_*$. Now let $T_+$ be the elements of $T$ 
with $\langle x_i - \hat{\mu}, v \rangle > 0$ and define $T_-$ similarly. 
We have
\begin{align}
\sum_{i \in T} |\langle x_i - \mu, v \rangle|
 &= \sum_{i \in T_+} \langle x_i - \mu, v \rangle - \sum_{i \in T_-} \langle x_i - \mu, v \rangle \\
 &= -\sum_{i \in S \backslash T_+} \langle x_i - \mu, v \rangle + \sum_{i \in S \backslash T_-} \langle x_i - \mu, v \rangle \\
 &\leq \Big\|\sum_{i \in S \backslash T_+} (x_i - \mu)\Big\|\|v\|_* + \Big\|\sum_{i \in S \backslash T_-} (x_i - \mu)\Big\|\|v\|_* \\
 &\leq 2|S|\sigmabest(\epsilon)\|v\|_*,
\end{align}
which yields the desired result.
\end{prooff}

%% file: unitary-proof.tex
\section{Fast Algorithm for $\ell_2$ Mean Estimation}
\label{sec:unitary-proof}

Recall that in Section~\ref{sec:main-l2} we wanted to solve the problem 
\begin{equation}
\min_W \|X - XW\|_2 \text{ subject to } 0 \leq W_{ij} \leq \frac{1}{\alpha n}, \bi^{\top}W = \bi^{\top}.
\end{equation}
An examination of the proof of Proposition~\ref{prop:recovery-l2} reveals that it 
actually suffices to replace the constraint $0 \leq W_{ij} \leq \frac{1}{\alpha n}$ with the 
Frobenius norm constraint $\|W\|_F^2 \leq \frac{1}{\alpha}$. We will show how to solve this 
modified problem efficiently (via a singular value decomposition and linear-time 
post-processing).
For $r = \frac{1}{\alpha}$, we want to solve
\begin{equation}
\min_W \|X - XW\|_2 \text{ subject to } \bi^{\top}W = \bi^{\top}, \|W\|_F^2 \leq r.
\end{equation}
Let $P = \frac{1}{n} \bi\bi^{\top}$. Then given any $W$, we can pass to the matrix 
$W' = P + (I-P)W$, which will always satisfy $\bi^{\top}W' = \bi^{\top}$. 
Therefore, it suffices to minimize 
\begin{equation}
\|X-XW'\|_2^2 = \|X-X(P + (I-P)W)\|_2^2 = \|X(I-P) - X(I-P)W\|_2^2
\end{equation}
with a constraint on
\begin{equation}
\|W'\|_F^2 = \|P + (I-P)W\|_F^2 = \|P\|_F^2 + \|(I-P)W\|_F^2. 
\end{equation}
Now, at the optimum we have $W = (I-P)W$, and for all $W$ we have 
$\|(I-P)W\|_F^2 \leq \|W\|_F^2$, so we can replace $\|(I-P)W\|_F^2$ with $\|W\|_F^2$ without 
changing the optimization. In addition, $P$ is a constant and $\|P\|_F^2 = 1$, 
so the constraint boils down to $\|W\|_F^2 \leq r-1$.
Letting $X' = X(I-P)$, it therefore suffices to solve
\begin{equation}
\min_W \|X' - X'W\|_2^2 \text{ subject to } \|W\|_F^2 \leq r-1.
\end{equation}
This problem is unitarily invariant and so can be solved efficiently via an SVD. 
In particular let $X' = U \Lambda V^{\top}$. The optimal $W$ lies in the span of 
$V^{\top}$, so after an appropriate change of variables it suffices to minimize 
$\|\Lambda - \Lambda W\|_2^2$ subject to $\|W\|_F^2 \leq r-1$. 
We can also check that the optimal $W$ is diagonal in this basis, 
so we need only solve the vector minimization 
\begin{equation}
\min_w \max_j \lambda_j^2 (1-w_j)^2 \text{ subject to } \sum_j w_j^2 \leq r-1,
\end{equation}
which can be done in linear time.

%% file: invariant-proof.tex
\vskip -0.3in
\phantom{x}
\section{Proof of Lemma~\ref{lem:invariant}}
\label{sec:invariant-proof}

Assuming that (i) and (ii) hold prior to line~\ref{line:outlier-begin} of the algorithm, we 
need to show that they continue to hold.

By assumption, we have $\sum_{i \in S \cap \sA} c_i\tau_i \leq \alpha a$, while 
$\sum_{i \in \sA} c_i\tau_i \geq 4a$. But note that the amount that 
$\sum_{i \in S} c_i$ decreases in line~\ref{line:outlier-begin} is proportional to $\sum_{i \in S \cap \sA} c_i\tau_i$, 
while the amount that $\sum_{i=1}^n c_i$ decreases is proportional to $\sum_{i \in \sA} c_i\tau_i$. 
Therefore, the former decreases at most $\frac{\alpha}{4}$ times as fast as the latter, 
meaning that (i) is preserved.

Since (i) is preserved, we have 
$\sum_{i \in S} (1-c_i) \leq \frac{\alpha}{4}\big(\sum_{i \in S} (1-c_i) + \sum_{i \not\in S} (1-c_i)\big)$. 
Re-arranging yields $\sum_{i \in S} (1-c_i) \leq \frac{\alpha}{4-\alpha} \sum_{i \not\in S} (1-c_i) \leq \frac{\alpha(1-\alpha)}{4-\alpha}n$. 
In particular, 
$\#\{i \mid c_i \leq \frac{1}{2}\} \leq \frac{2\alpha(1-\alpha)}{4-\alpha}n$, 
so at most $\frac{2\alpha(1-\alpha)}{4-\alpha}n$ elements of $S$ have 
been removed from $\sA$ in total. 
This implies $|S \cap \sA| \geq \frac{\alpha(2+\alpha)}{4-\alpha} n$, and 
so (ii) is preserved as well. \qed

%% file: concentration-cov-proof.tex
\section{Proof of Proposition~\ref{prop:concentration-cov}}
\label{sec:concentration-cov-proof}

This is a corollary of Proposition B.1 of \citet{charikar2017learning}, 
which states that given a distribution with covariance $\Sigma_0$, where 
$\Sigma_0 \preceq \sigma_0^2 I$, with probability 
$1 - \exp(-\epsilon^2 n/16)$ it is possible to find a set $T$ of $(1-\epsilon)n$ points 
such that $\frac{1}{|T|} \sum_{i \in T} (x_i - \mu)(x_i - \mu)^{\top} \preceq (\sigma_0')^2 I$, 
where $\sigma_0'$ and $\sigma_0$ have the same relation as $\sigma'$ and $\sigma$ above.
Applying this proposition directly to $p$ will not give us our result, so we instead 
apply it to a transformed version of $p$.

In particular, let $\Sigma$ be the covariance of $p$, and given a sample $x_i \sim p$, 
let $y_i = \Sigma^{-1/2}(x_i - \mu)$.\footnote{In the case that $\Sigma$ is not invertible, 
we need to add a small multiple of the identity matrix before inverting, but we omit this 
part of the argument for brevity.} Then the $y_i$ have identity covariance and mean $0$, 
and so Proposition B.1 of \citet{charikar2017learning} implies that we can find 
$(1-\epsilon)n$ of the $y_i$ such that
%\begin{equation}
$\frac{1}{|T|} \sum_{i \in T} y_iy_i^{\top} \preceq \frac{4}{\epsilon}\p{1 + \frac{d}{(1-\epsilon)n}} I$.
%\end{equation}
In terms of the $x_i$, after multiplying by $\Sigma^{1/2}$ on the left and right, 
we obtain
\begin{equation}
\frac{1}{|T|} \sum_{i \in T} (x_i-\mu)(x_i-\mu)^{\top} \preceq \frac{4}{\epsilon}\p{1 + \frac{d}{(1-\epsilon)n}} \Sigma,
\end{equation}
which implies that 
\begin{align}
\frac{1}{|T|} \sum_{i \in T} \langle x_i - \mu, v \rangle^2 
 &\leq \frac{4}{\epsilon}\p{1 + \frac{d}{(1-\epsilon)n}} v^{\top}\Sigma v \\
 &= \frac{4}{\epsilon}\p{1 + \frac{d}{(1-\epsilon)n}} \bE_{x \sim p}[\langle x - \mu, v \rangle^2] \\
 &\leq \frac{4}{\epsilon}\p{1 + \frac{d}{(1-\epsilon)n}} \sigma^2 \|v\|_*^2 
  = (\sigma')^2 \|v\|_*^2,
\end{align}
as was to be shown.

%% file: main.bbl
\begin{thebibliography}{29}
\providecommand{\natexlab}[1]{#1}
\providecommand{\url}[1]{\texttt{#1}}
\expandafter\ifx\csname urlstyle\endcsname\relax
  \providecommand{\doi}[1]{doi: #1}\else
  \providecommand{\doi}{doi: \begingroup \urlstyle{rm}\Url}\fi

\bibitem[Balakrishnan et~al.(2017)Balakrishnan, Du, Li, and
  Singh]{balakrishnan2017sparse}
S.~Balakrishnan, S.~S. Du, J.~Li, and A.~Singh.
\newblock Computationally efficient robust sparse estimation in high
  dimensions.
\newblock In \emph{Conference on Learning Theory (COLT)}, pages 169--212, 2017.

\bibitem[Batson et~al.(2012)Batson, Spielman, and Srivastava]{batson2012twice}
J.~Batson, D.~A. Spielman, and N.~Srivastava.
\newblock Twice-{R}amanujan sparsifiers.
\newblock \emph{SIAM Journal on Computing}, 41\penalty0 (6):\penalty0
  1704--1721, 2012.

\bibitem[Charikar et~al.(2017)Charikar, Steinhardt, and
  Valiant]{charikar2017learning}
M.~Charikar, J.~Steinhardt, and G.~Valiant.
\newblock Learning from untrusted data.
\newblock In \emph{Symposium on Theory of Computing (STOC)}, 2017.

\bibitem[Decelle et~al.(2011)Decelle, Krzakala, Moore, and
  Zdeborov{\'a}]{decelle2011asymptotic}
A.~Decelle, F.~Krzakala, C.~Moore, and L.~Zdeborov{\'a}.
\newblock Asymptotic analysis of the stochastic block model for modular
  networks and its algorithmic applications.
\newblock \emph{Physical Review E}, 84\penalty0 (6), 2011.

\bibitem[Diakonikolas et~al.(2016{\natexlab{a}})Diakonikolas, Kamath, Kane, Li,
  Moitra, and Stewart]{diakonikolas2016robust}
I.~Diakonikolas, G.~Kamath, D.~Kane, J.~Li, A.~Moitra, and A.~Stewart.
\newblock Robust estimators in high dimensions without the computational
  intractability.
\newblock In \emph{Foundations of Computer Science (FOCS)}, 2016{\natexlab{a}}.

\bibitem[Diakonikolas et~al.(2016{\natexlab{b}})Diakonikolas, Kane, and
  Stewart]{diakonikolas2016bayes}
I.~Diakonikolas, D.~Kane, and A.~Stewart.
\newblock Robust learning of fixed-structure {B}ayesian networks.
\newblock \emph{arXiv}, 2016{\natexlab{b}}.

\bibitem[Diakonikolas et~al.(2016{\natexlab{c}})Diakonikolas, Kane, and
  Stewart]{diakonikolas2016statistical}
I.~Diakonikolas, D.~M. Kane, and A.~Stewart.
\newblock Statistical query lower bounds for robust estimation of
  high-dimensional {G}aussians and {G}aussian mixtures.
\newblock \emph{arXiv}, 2016{\natexlab{c}}.

\bibitem[Diakonikolas et~al.(2017{\natexlab{a}})Diakonikolas, Kamath, Kane, Li,
  Moitra, and Stewart]{diakonikolas2017practical}
I.~Diakonikolas, G.~Kamath, D.~Kane, J.~Li, A.~Moitra, and A.~Stewart.
\newblock Being robust (in high dimensions) can be practical.
\newblock \emph{arXiv}, 2017{\natexlab{a}}.

\bibitem[Diakonikolas et~al.(2017{\natexlab{b}})Diakonikolas, Kamath, Kane, Li,
  Moitra, and Stewart]{diakonikolas2017robustly}
I.~Diakonikolas, G.~Kamath, D.~M. Kane, J.~Li, A.~Moitra, and A.~Stewart.
\newblock Robustly learning a {G}aussian: Getting optimal error, efficiently.
\newblock \emph{arXiv}, 2017{\natexlab{b}}.

\bibitem[Diakonikolas et~al.(2017{\natexlab{c}})Diakonikolas, Kane, and
  Stewart]{diakonikolas2017learning}
I.~Diakonikolas, D.~M. Kane, and A.~Stewart.
\newblock Learning geometric concepts with nasty noise.
\newblock \emph{arXiv}, 2017{\natexlab{c}}.

\bibitem[Gu{\'e}don and Vershynin(2014)]{guedon2014community}
O.~Gu{\'e}don and R.~Vershynin.
\newblock Community detection in sparse networks via {G}rothendieck's
  inequality.
\newblock \emph{arXiv}, 2014.

\bibitem[Haagerup(1981)]{haagerup1981best}
U.~Haagerup.
\newblock The best constants in the khintchine inequality.
\newblock \emph{Studia Mathematica}, 70\penalty0 (3):\penalty0 231--283, 1981.

\bibitem[Kane et~al.(2017)Kane, Karmalkar, and Price]{kane2017robust}
D.~Kane, S.~Karmalkar, and E.~Price.
\newblock Robust polynomial regression up to the information theoretic limit.
\newblock \emph{arXiv}, 2017.

\bibitem[Khintchine(1923)]{khintchine1923uber}
A.~Khintchine.
\newblock {\"U}ber dyadische br{\"u}che.
\newblock \emph{Mathematische Zeitschrift}, 18:\penalty0 109--116, 1923.

\bibitem[Klivans et~al.(2009)Klivans, Long, and Servedio]{klivans2009learning}
A.~R. Klivans, P.~M. Long, and R.~A. Servedio.
\newblock Learning halfspaces with malicious noise.
\newblock \emph{Journal of Machine Learning Research (JMLR)}, 10:\penalty0
  2715--2740, 2009.

\bibitem[Kothari and Steinhardt(2017)]{kothari2017agnostic}
P.~Kothari and J.~Steinhardt.
\newblock Better agnostic clustering via tensor norms.
\newblock \emph{arXiv}, 2017.

\bibitem[Lai et~al.(2016)Lai, Rao, and Vempala]{lai2016agnostic}
K.~A. Lai, A.~B. Rao, and S.~Vempala.
\newblock Agnostic estimation of mean and covariance.
\newblock In \emph{Foundations of Computer Science (FOCS)}, 2016.

\bibitem[Le et~al.(2015)Le, Levina, and Vershynin]{le2015concentration}
C.~M. Le, E.~Levina, and R.~Vershynin.
\newblock Concentration and regularization of random graphs.
\newblock \emph{arXiv}, 2015.

\bibitem[Li(2017)]{li2017sparse}
J.~Li.
\newblock Robust sparse estimation tasks in high dimensions.
\newblock \emph{arXiv}, 2017.

\bibitem[Massouli{\'e}(2014)]{massoulie2014community}
L.~Massouli{\'e}.
\newblock Community detection thresholds and the weak {R}amanujan property.
\newblock In \emph{Symposium on Theory of Computing (STOC)}, pages 694--703,
  2014.

\bibitem[Meister and Valiant(2017)]{meister2017data}
M.~Meister and G.~Valiant.
\newblock A data prism: Semi-verified learning in the small-alpha regime.
\newblock \emph{arXiv}, 2017.

\bibitem[Mossel et~al.(2013)Mossel, Neeman, and Sly]{mossel2013proof}
E.~Mossel, J.~Neeman, and A.~Sly.
\newblock A proof of the block model threshold conjecture.
\newblock \emph{arXiv}, 2013.

\bibitem[Nesterov(1998)]{nesterov1998semidefinite}
Y.~Nesterov.
\newblock Semidefinite relaxation and nonconvex quadratic optimization.
\newblock \emph{Optimization methods and software}, 9:\penalty0 141--160, 1998.

\bibitem[Rebrova and Tikhomirov(2015)]{rebrova2015coverings}
E.~Rebrova and K.~Tikhomirov.
\newblock Coverings of random ellipsoids, and invertibility of matrices with
  iid heavy-tailed entries.
\newblock \emph{arXiv}, 2015.

\bibitem[Rebrova and Vershynin(2016)]{rebrova2016norms}
E.~Rebrova and R.~Vershynin.
\newblock Norms of random matrices: local and global problems.
\newblock \emph{arXiv}, 2016.

\bibitem[Shalev-Shwartz(2007)]{shalev07online}
S.~Shalev-Shwartz.
\newblock \emph{Online Learning: Theory, Algorithms, and Applications}.
\newblock PhD thesis, The Hebrew University of Jerusalem, 2007.

\bibitem[Steinhardt(2017)]{steinhardt2017clique}
J.~Steinhardt.
\newblock Does robustness imply tractability? {A} lower bound for planted
  clique in the semi-random model.
\newblock \emph{arXiv}, 2017.

\bibitem[Steinhardt et~al.(2016)Steinhardt, Valiant, and
  Charikar]{steinhardt2016avoiding}
J.~Steinhardt, G.~Valiant, and M.~Charikar.
\newblock Avoiding imposters and delinquents: Adversarial crowdsourcing and
  peer prediction.
\newblock In \emph{Advances in Neural Information Processing Systems (NIPS)},
  2016.

\bibitem[Xu et~al.(2010)Xu, Caramanis, and Mannor]{xu2010principal}
H.~Xu, C.~Caramanis, and S.~Mannor.
\newblock Principal component analysis with contaminated data: The high
  dimensional case.
\newblock \emph{arXiv}, 2010.

\end{thebibliography}
